\documentclass[letterpaper]{article} 
\usepackage{aaai25}  
\usepackage{times}  
\usepackage{helvet}  
\usepackage{courier}  
\usepackage[hyphens]{url}  
\usepackage{graphicx} 
\usepackage{natbib}  
\usepackage{caption} 
\usepackage{algorithm}
\usepackage{algorithmic}
\usepackage{amsmath,amssymb}    
\usepackage{multirow}
\usepackage{graphicx}
\usepackage{newfloat}
\usepackage{listings}
\DeclareCaptionStyle{ruled}{labelfont=normalfont,labelsep=colon,strut=off} 
\floatstyle{ruled}
\newfloat{listing}{tb}{lst}{}
\floatname{listing}{Listing}

\title{TVG: A Training-free Transition Video Generation Method \\ with Diffusion Models}
\author{
    Rui Zhang\textsuperscript{\rm 1,2,3}\thanks{This work was done during an internship at Sobey Media Intelligence Laboratory}
    Yaosen Chen\textsuperscript{\rm 1,4}\thanks{Co-corresponding authors}, 
    Yuegen Liu\textsuperscript{\rm 1}, 
    Wei Wang\textsuperscript{\rm 1}, 
    Xuming Wen\textsuperscript{\rm 1}, 
    Hongxia Wang\textsuperscript{\rm 2,3}\thanks{Co-corresponding authors}
}
\affiliations{
    \textsuperscript{\rm 1}Sobey Media Intelligence Laboratory, Chengdu, China\\
    \textsuperscript{\rm 2}School of Cyber Science and Engineering, Sichuan University, Chengdu, China\\
    \textsuperscript{\rm 3}Key Laboratory of Data Protection and Intelligent Management (Sichuan University), Ministry of Education, China\\
    \textsuperscript{\rm 4}University of Electronic Science and Technology of China, Chengdu, Sichuan, 611731 China\\
}

\usepackage{bibentry}

\begin{document}
\nocopyright
\maketitle



\begin{abstract}
Transition videos play a crucial role in media production, enhancing the flow and coherence of visual narratives. Traditional methods like morphing often lack artistic appeal and require specialized skills, limiting their effectiveness. Recent advances in diffusion model-based video generation offer new possibilities for creating transitions but face challenges such as poor inter-frame relationship modeling and abrupt content changes. We propose a novel training-free Transition Video Generation (TVG) approach using video-level diffusion models that addresses these limitations without additional training. Our method leverages Gaussian Process Regression ($\mathcal{GPR}$) to model latent representations, ensuring smooth and dynamic transitions between frames. Additionally, we introduce interpolation-based conditional controls and a Frequency-aware Bidirectional Fusion (FBiF) architecture to enhance temporal control and transition reliability. Evaluations of benchmark datasets and custom image pairs demonstrate the effectiveness of our approach in generating high-quality smooth transition videos. The code are provided in \url{https://sobeymil.github.io/tvg.com}.
\end{abstract}
\section{Introduction}
\label{Introduction}

Transition videos have been fundamental in media production, traditionally using techniques such as morphing~\cite{DBLP:journals/vc/Wolberg98} to seamlessly connect segments, thus improving flow, visual appeal, and narrative coherence. With video emerging as the dominant medium on platforms such as TikTok and Kuaishou, the demand for innovative content formats such as vlogs and short films has surged. This trend underscores the critical need for effective and engaging transitions. However, traditional transition techniques often lack artistic appeal, fail to fully engage viewers, and require specialized production skills, posing challenges for creators.

\begin{figure}[t]
    \centering
    \includegraphics[width=\columnwidth]{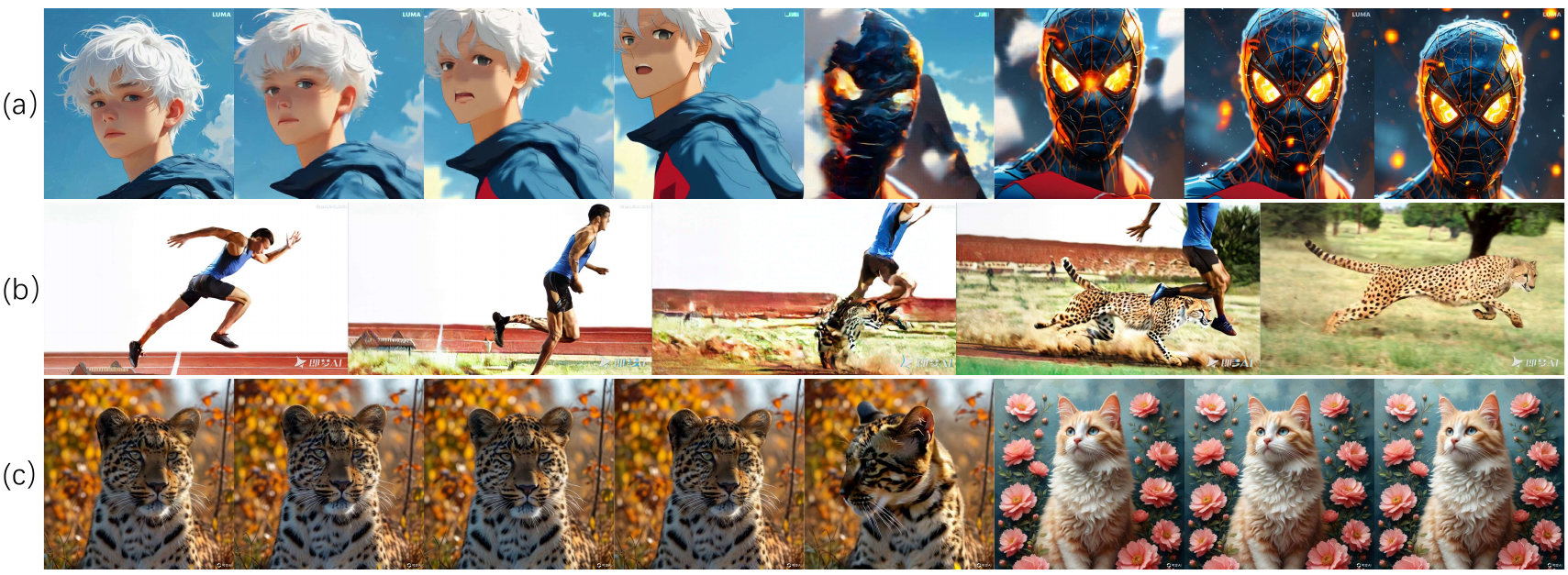}
    \caption{Failed results from some commercial products: (a) LUMA AI, (b) Jimeng AI, (c) Kling AI.}
    \label{fig:failed_example}
\end{figure}

Recent advancements in diffusion model-based image~\cite{DBLP:conf/iclr/SongME21, rombach2022high} and video generation~\cite{liu2024sora} have introduced novel methods for creating transition videos by generating intermediate frames between segments. Although these approaches offer more creative and accessible transition generation, they face limitations, especially when there is a significant disparity between frames. Existing techniques, including both open-source models~\cite{Zhang_2024_CVPR, DBLP:conf/iclr/Chen0ZZMY0L0024, xing2023dynamicrafter} and commercial products, often produce suboptimal transitions, leading to abrupt changes, simple fades, or even the absence of transitions, as shown in Figure~\ref{fig:failed_example}. Additionally, current video generation models are typically limited to producing short clips, making effective transition generation crucial for maintaining continuity in longer videos~\cite{yuan2024mora}. The inability to generate consistent, high-quality transitions could significantly hinder the broader adoption of video generation technologies.

Several factors contribute to these challenges. First, some models~\cite{Zhang_2024_CVPR, DBLP:conf/iclr/Chen0ZZMY0L0024} extend and optimize image-level diffusion models, which often lack robust inter-frame relationship modeling, particularly when the primary subjects in the frames differ significantly. Although these models generally avoid abrupt transitions, they often fail to produce dynamic results, leading to outputs with simple fading effects or disruptive artifacts. Second, while video-level diffusion models~\cite{xing2023dynamicrafter} attempt to establish inter-frame relationships through spatio-temporal attention, these can be unstable due to issues like conditional image leakage~\cite{zhao2024identifying}, cross-attention leakage~\cite{yang2024dynamic}, and improper activation of attention mechanisms, resulting in abrupt transitions or anomalous content.

To address these challenges, we propose a novel, training-free \textbf{T}ransition \textbf{V}ideo \textbf{G}eneration~(TVG) approach based on video-level diffusion models. Unlike methods like SEINE~\cite{DBLP:conf/iclr/Chen0ZZMY0L0024}, which require full retraining, or DiffMorpher~\cite{Zhang_2024_CVPR}, which relies on LoRA-based~\cite{hu2022lora} fine-tuning, our approach generates video transitions without any additional training. While existing image-to-video pre-trained models demonstrate high-quality generative capabilities, they often lack effective inter-frame relationship control for transitions. Our approach avoids redundant retraining, reducing computational costs and enabling broader applicability to future, potentially more advanced video generation algorithms.

We implemented several optimizations based on DynamiCrafter~\cite{xing2023dynamicrafter}. Building on prior work~\cite{Zhang_2024_CVPR, jang2024spherical}, we introduce a novel conditional control approach using interpolation techniques. Specifically, we blend images within the conditional latent space, enhancing motion dynamics. Additionally, we utilize spherical linear interpolation (SLERP) for prompt embeddings, enabling precise control and fine-grained adjustments of each frame's content, addressing the lack of dynamic variation caused by conditional image leakage. We propose that the content between transition frames should adhere to a smooth stochastic process to balance dynamic motion with content continuity. To achieve this, we employ \textbf{G}aussian \textbf{P}rocess \textbf{R}egression~($\mathcal{GPR}$) to model the latent representations of the initial and final frames. The intermediate content is then mapped into this smooth feature space, improving transition quality by preventing abrupt changes while maintaining dynamic continuity. Furthermore, we observed that existing image-to-video models can produce vastly different transition videos depending on the content of the initial and final frames, even when simply swapping the two. To address this variability and enhance the reliability of transition generation, we introduced a \textbf{F}requence-aware \textbf{Bi}directional \textbf{F}usion~(FBiF) architecture. We evaluated our algorithm on existing benchmarks, including MorphBench~\cite{Zhang_2024_CVPR} and TC-Bench-I2V~\cite{DBLP:journals/corr/abs-2406-08656}, as well as on custom image pairs. The experimental results demonstrate the effectiveness of our proposed modules. Our contributions are summarized as follows:

\begin{enumerate}
    \item We propose a training-free transition video generation method based on \textbf{G}aussian \textbf{P}rocess \textbf{R}egression~($\mathcal{GPR}$) with video-level diffusion models.
    \item We introduce interpolation-based conditional controls to enhance temporal control and ensure smooth video transitions.
    \item We develop a \textbf{F}requence-aware \textbf{Bi}directional \textbf{F}usion~(FBiF) architecture to further improve the reliability of video transition generation.
\end{enumerate}

\section{Related work}
\label{related_work}

\noindent\textbf{Training-free Video Generation.} ControlVideo~\cite{zhang2023controlvideo}, adapted from ControlNet, improves video generation by introducing cross-frame interaction, interleaved-frame smoothing, and hierarchical sampling. MotionClone~\cite{ling2024motionclone} employs temporal attention in video inversion for motion representation and cloning, while VideoElevator~\cite{zhang2024videoelevator} decomposes sampling into temporal motion refinement and spatial quality enhancement. MVOC~\cite{wang2024mvoc} propose a multiple video object composition  method based on diffusion models with DDIM inversion features injection. These methods enhance video generation and editing capabilities by building on existing models, reducing the need for extensive retraining.


\noindent\textbf{Transition Video Generation.}
Traditional video transitions, such as fade, dissolve, wipe, or cut, are essential for seamless scene changes in video editing. Morphing~\cite{DBLP:journals/vc/Wolberg98} offers smooth transitions by finding pixel-level similarities and estimating offsets. Generative models~\cite{van2017neural} have advanced this by capturing semantic similarities through latent code interpolation, with applications in style transfer~\cite{chen2018gated} and object transfiguration~\cite{DBLP:conf/siggraph/SauerS022}. SEINE~\cite{DBLP:conf/iclr/Chen0ZZMY0L0024} introduced a random-mask video diffusion model for text-guided transitions between different scenes, while DiffMorpher~\cite{Zhang_2024_CVPR} uses LoRA parameter interpolation with image diffusion models for smooth semantic transitions. Despite these advances, achieving natural video transitions remains challenging.


\section{Preliminary}

\subsection{Diffusion Models}
Diffusion models~\cite{sohl2015deep,ho2020denoising,rombach2022high} are a class of probabilistic generative models which learns a data distribution by denoising noisy. The forward process of diffuses the data samples with a Markov process contains $T$ timesteps:
The forward process is gradually adding noise to the data sample $x_0$ to $x_t$ with a Markov process contains $T$ timesteps:
\begin{subequations}
\begin{equation} \label{eq:qxtx0}
	\begin{aligned}
		q(\boldsymbol{x}_{t}|\boldsymbol{x}_{t-1})= \mathcal{N}\bigl(\boldsymbol{x}_{t};\sqrt{1-\beta_t}\boldsymbol{x}_{t-1},\beta_t\boldsymbol{I}),	
	\end{aligned}
\end{equation}
\begin{equation} \label{eq:qxtx1}
	\begin{aligned}
		q( \boldsymbol{x}_t|\boldsymbol{x}_0)=  \mathcal{N}\bigl(\boldsymbol{x}_{t}; \sqrt{\overline{\alpha}_t}\boldsymbol{x}_0, (1- \overline{\alpha}_t)I\bigl),
	\end{aligned}
\end{equation}
\end{subequations}
where $\beta_t$ is a predefined variance schedule, $\overline{\alpha}_t=\prod_{s=1}^t\alpha_s$, is derived from the variance noise schedule, and $\alpha_t=1-\beta_t$. The reverse process begins from the noise and transitions towards the original data $q(\boldsymbol{x}_0)$.  
The  reverse denoising process obtains less noisy data $x_{t-1}$ from the noisy input $x_t$ at each timestep:
\begin{equation} \label{eq:qxtx2}
	\begin{aligned}
		p_{{{\theta}_{l}}}(\boldsymbol{x}_{t-1}|\boldsymbol{x}_{t})= \mathcal{N}\bigl(\boldsymbol{x}_{t};\frac{1}{\sqrt{\alpha_{t}}}(\boldsymbol{x}_{t}-\frac{1-\alpha_{t}}{\sqrt{1-\bar{\alpha}}_{t}}\epsilon_{{{\theta}_{l}}}(\boldsymbol{x}_{t},t)),\sigma^{2}_{t})	 
	\end{aligned}
\end{equation}
where $\epsilon_{{\theta}_{l}}(\boldsymbol{x}_t, t)$ is the denoising function to estimate the noise added at each step and ${{\theta}_{l}}$ is the learnable network parameters.

\subsection{Diffusion Models for Transition Video Generation}

Given two input images, $x_0$ and $x_{S}$, the goal of transition video generation is to produce a video sequence $X_{t} = \{x_i\}_{i=1}^{S}$, starting with $x_0$ and ending with $x_{S}$. The intermediate frames, $x_1$ to $x_{S-1}$, should exhibit smooth transitions that blend the characteristics of both input images, forming a coherent sequence. This process can be formalized as a conditional distribution $p_{{\theta}_{l}}(X_{t} \mid x_0, x_{S})$, modeled by a conditional diffusion process.

In video generation, Latent Diffusion Models (LDMs)~\cite{rombach2022high,zhou2022magicvideo,an2023latent} are commonly used to reduce computational complexity by fitting the conditional distribution within the latent space, where denoising occurs. The initial and final frames, $x_0$ and $x_S$, anchor the sequence, with intermediate frames initialized by adding Gaussian noise. These frames are encoded into a latent representation $z_t \in \mathbb{R}^{S \times C \times H \times W}$, where $S$, $C$, $H$, and $W$ represent the number of frames, channels, height, and width, respectively. A backward denoising process, such as DDIM~\cite{DBLP:conf/iclr/SongME21}, is applied to obtain the clean latent representation $z_0$, which is then decoded into the video sequence $X_t$. The reverse process typically incorporates additional denoising conditions to enhance the quality of the generated video, such as text prompt latents, $c_{\text{txt}}$, and the latent of the initial frame, $c_{\text{img}}$.

Our base model, DynamiCrafter~\cite{xing2023dynamicrafter}, further leverages FPS as a denoising condition. It uses CLIP~\cite{Radford2021LearningTV} to encode text prompts and the initial frame, while employing a frame-wise Variational Auto-Encoder (VAE) architecture for the video encoder and decoder. The denoising process is carried out using a U-Net architecture.

\subsection{Gaussian Process}



A Gaussian process ($\mathcal{GP}$)~\cite{williams2006gaussian} is a collection of random variables, any finite number of which have a joint Gaussian distribution. Consider a data set with inputs and targets consisting of $N$ data-points  as $\{x_i, y_i \}_{i=1}^{N}$, where the inputs are denoted by $\boldsymbol{X}$ =$\{x_{1}, . . . , x_{N}\}$ , and targets by $\boldsymbol{Y}$ =$\{y_{1}, . . . , y_{N}\}$.
The goal of $\mathcal{GP}$ is to learn an unknown function $f$ that maps elements from input space to a target space, which can provide predictive distributions on test points  $\boldsymbol{X^*}$ =$\{x^*_{1}, . . . , x^*_{M}\}$. Assuming the GP prior on $f(x)$ with some additive Gaussian white noise with variance $\sigma^{2}$:
\begin{equation} \label{eq:gps01}
	\begin{aligned}
	y_i = f(x_i) +\epsilon_{i}, \epsilon_{i} \sim \mathcal{N}(0,\sigma^{2})
	\end{aligned}
\end{equation}

A Gaussian process regression~($\mathcal{GPR}$)  formulates the functional form of $f$ by drawing random variables from a multi-variate normal distribution given by
$[f(x_1),f(x_2),...,f(x_n)]\sim\mathcal{N}(\mu,\boldsymbol{K}_{\boldsymbol{X},\boldsymbol{X}})$, with the mean of $\mu$ and the covariance matrix of $\boldsymbol{K}_{\boldsymbol{X},\boldsymbol{X}}$. $\mu_{i}=\mu(x_{i})$, where $\mu(\cdot)$ is a mean function of the $\mathcal{GP}$; $(\boldsymbol{K}_{\boldsymbol{X},\boldsymbol{X}})_{ij} = k(x_i, x_j)$, where $k(\cdot)$ is a kernel function of the $\mathcal{GP}$.
The conditional distribution at any unseen points $\boldsymbol{X}^{*}$ is given by:
\begin{equation} \label{eq:gps02}
	\begin{aligned}
		\boldsymbol{f^{*}}|\boldsymbol{X^{*}},\boldsymbol{X},\boldsymbol{Y}\sim\mathcal{N}(\mu_{\boldsymbol{f^{*}}},\Sigma_{\boldsymbol{f^{*}}})
	\end{aligned}
\end{equation}
where 
\begin{equation} \label{eq:gps03}
	\begin{aligned}
		\mu_{\boldsymbol{f^{*}}}=\mu\boldsymbol{X^{*}}+\boldsymbol{K}_{\boldsymbol{X^*},\boldsymbol{X}}[\boldsymbol{K}_{\boldsymbol{X},\boldsymbol{X}}+\sigma^{2}\boldsymbol{I}]^{-1}\boldsymbol{Y}
	\end{aligned}
\end{equation}
\begin{equation} \label{eq:gps04}
	\begin{aligned}
		\Sigma_{\boldsymbol{f^{*}}}=\boldsymbol{K}_{\boldsymbol{X^*},\boldsymbol{X^*}}-\boldsymbol{K}_{\boldsymbol{X^{*}},\boldsymbol{X}}[\boldsymbol{K}_{\boldsymbol{X},\boldsymbol{X}}+\sigma^{2}\boldsymbol{I}]^{-1}\boldsymbol{K}_{\boldsymbol{X},\boldsymbol{X^*}}
	\end{aligned}
\end{equation}

\begin{figure*}[t]
    \centering
    \includegraphics[width=2\columnwidth]{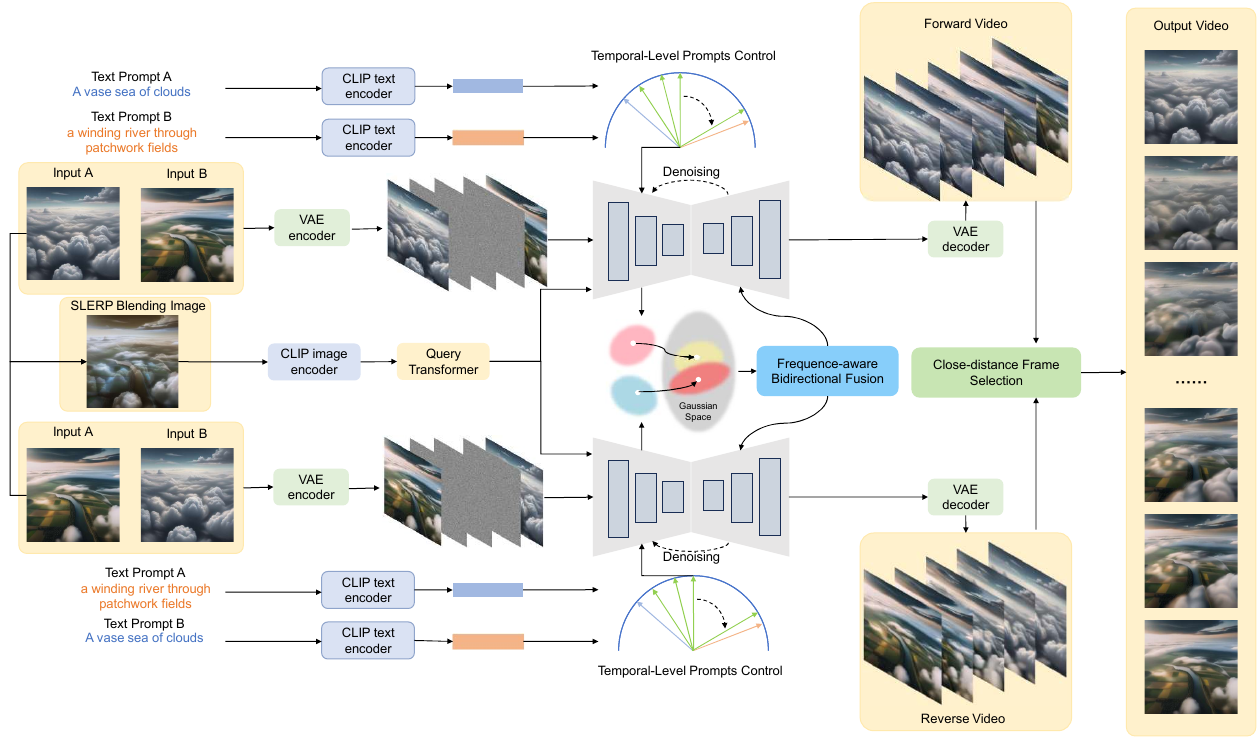}
    \caption{Illustration of Our Proposed Training-free TVG Method.}
    \label{fig:framework}
\end{figure*}

\section{Method}

Our framework is based on DynamiCrafter~\cite{xing2023dynamicrafter}, a state-of-the-art image-to-video generative model. However, DynamiCrafter faces challenges in video transition tasks, particularly when there are significant differences between the initial and final frames, often resulting in transitions that overly depend on the starting image and appear abrupt. To address these issues, we propose a training-free TVG method, as shown in Figure~\ref{fig:framework}.

We optimized the pre-trained model during inference in three key areas. First, we refined both conditional images and prompts to enhance the controllability of the video generation process and reduce conditional image leakage. Second, we integrated Gaussian Process Regression ($\mathcal{GPR}$) into the latent space to enforce temporal consistency between frames, improving inter-frame coherence and preventing abrupt transitions. Lastly, we introduced a Frequence-aware Bidirectional Fusion~(FBiF) structure, which combines bidirectional generation with frequency-domain feature fusion. To ensure the final output's visual quality, we apply close-distance frame selection at the final stage.

\subsection{Interpolation-based Conditional Controls}

In our experiments, we found that DynamiCrafter often produced abrupt transitions when using input images with significantly different content, a result of conditional image leakage. To mitigate this, we applied linear interpolation to blend the two input frames $x_0$ and $x_S$:

\begin{equation}
    \begin{aligned}
        x_b = \beta \cdot x_0 + (1-\beta) \cdot x_S.
    \end{aligned}
\end{equation}
Setting $\beta$ to 0.5 balances information from both frames, creating a blended image $x_b$ that integrates details from both inputs. Other values would favor one frame over the other, which is not suitable for our video transition task. This image is then processed by CLIP to extract feature embeddings, serving as the condition $c_{img}$. This approach leverages the pre-trained model without requiring additional training. In contrast, separate extraction and fusion of features using CLIP can create a semantic gap, complicating their integration in diffusion models and often leading to errors. Addressing this gap usually requires retraining a feature projection module, as done in StoryDiffusion~\cite{zhou2024storydiffusion}, but our method avoids this by fully utilizing the pre-trained model.

Traditional methods typically employ a single prompt for the entire video, limiting frame-wise control. To improve this, inspired by~\cite{Zhang_2024_CVPR, jang2024spherical}, we propose Temporal-Level Prompts Control. We begin by obtaining prompts $t_0$ and $t_S$ for the initial and final frames, then use CLIP to extract their semantic features, $F^0_{\text{txt}}$ and $F^S_{\text{txt}}$. These features are interpolated using spherical linear interpolation (SLERP).

\begin{equation}
    \begin{aligned}
        F^s_{\text{txt}} = \frac{\sin{((1-\alpha)\theta)}}{\sin{\theta}} \cdot F^0_{\text{txt}} + \frac{\sin{(\alpha\theta)}}{\sin{\theta}} \cdot F^S_{\text{txt}},
    \end{aligned}
\end{equation}
where $\alpha \in [0, 1]$ is the interpolation parameter, and $\theta = \cos^{-1}\left(\frac{F^0_{\text{txt}} \cdot F^S_{\text{txt}}}{\|F^0_{\text{txt}}\| \|F^S_{\text{txt}}\|}\right)$ is the angle between the two vectors. We sample $S$ weights uniformly between 0.9 and 0.1, corresponding to the temporal progression from the initial to the final frame, creating $S$ interpolated semantic features. These features form $c_{\text{txt}} \in \mathbb{R}^{S \times L \times D}$, ensuring smooth transitions and minimizing semantic gaps, allowing for seamless integration into the diffusion model and reducing the likelihood of abrupt content changes in the video.

\subsection{Latent Space Gaussian Process Regression}

The denoising U-Net employs temporal-level and spatial-level Transformer modules, integrating self-attention and cross-attention to capture spatial relationships within frames and temporal relationships between frames. However, the spatial modules overlook inter-frame dependencies, while the temporal modules impose only limited inter-frame constraints through attention mechanisms. This results in poor controllability, leading to error propagation across frames and degrading the quality of transition videos.

To address these issues, we incorporate $\mathcal{GPR}$ into the latent space of the U-Net within video generation models. This integration with attention modules explicitly constrains inter-frame relationships, enhancing the overall video quality.

Given a temporal-level latent vector $z \in \mathbb{R}^{S \times N \times P}$, where $N$ represents the vector size and $P$ denotes the number of channels, we compute the attention mechanism with $\mathcal{GPR}$ for both temporal and spatial attention as follows:

\begin{equation}
    \begin{aligned}
    z^{\prime} = \text{Attention}(z) + \gamma \cdot z + (1-\gamma) \cdot \mathcal{GPR}(z),
    \end{aligned}
\end{equation}
where $\gamma \in [0, 1]$ serves as a balancing factor, controlling the trade-off between different feature contributions. We hypothesize that the transition content between two frames in a video can be modeled as a smooth stochastic process. The objective is to derive a distribution function for this process, with the first frame as the observation input and the last frame as the prediction target.

In the latent space, the features of the first and last frames are extracted from the input images, while intermediate frames are generated by denoising Gaussian noise. This allows us to construct a stochastic process distribution function based on the initial and final frames, predicting intermediate frames to ensure coherent transitions that accurately reflect the relationship between the start and end frames.

To implement this, we leverage $\mathcal{GPR}$ to model the distribution function, utilizing its capability to flexibly capture the underlying data distribution and smoothly interpolate between frames while maintaining consistency with the input frames. Specifically, the first frame of $z$ is defined as $X \in \mathbb{R}^{N \times P},$ where $X$ consists of $N$ data points serving as input to $\mathcal{GPR}$, as outlined in the Preliminary section. Similarly, the last frame of $z$ is defined as $Y \in \mathbb{R}^{N \times P},$ representing the target with $N$ data points. We employ the Radial Basis Function as the kernel $k(\cdot)$ to formulate the distribution function $\boldsymbol{f^{*}}.$ The remaining frames of $z$ are used as the input $\boldsymbol{X^{*}}$ to the function, as shown in Equation~\ref{eq:gps02}. The mean $\mu_{\boldsymbol{f^{*}}}$ obtained from the function, as shown in Equation~\ref{eq:gps03}, is then used to refine the corresponding frame features.

This approach effectively maps features into a Gaussian space, aligning inter-frame relationships with a stationary random distribution, thereby enhancing both correlation and smoothness between frames. However, due to inherent differences between the original model's feature distribution and the Gaussian distribution, a complete substitution of the original features would necessitate further training to achieve optimal performance. To reduce computational costs and avoid additional training, we integrate these features as a penalty term within the original attention mechanism, rather than fully replacing the original features. Despite this conservative approach, it significantly enhances inter-frame information exchange and smooths transitions between frames, while maintaining compatibility with the feature distribution of the pre-trained model.

\subsection{Frequence-aware Bidirectional Fusion}
\begin{figure}[htbp]
    \centering
    \includegraphics[width=\columnwidth]{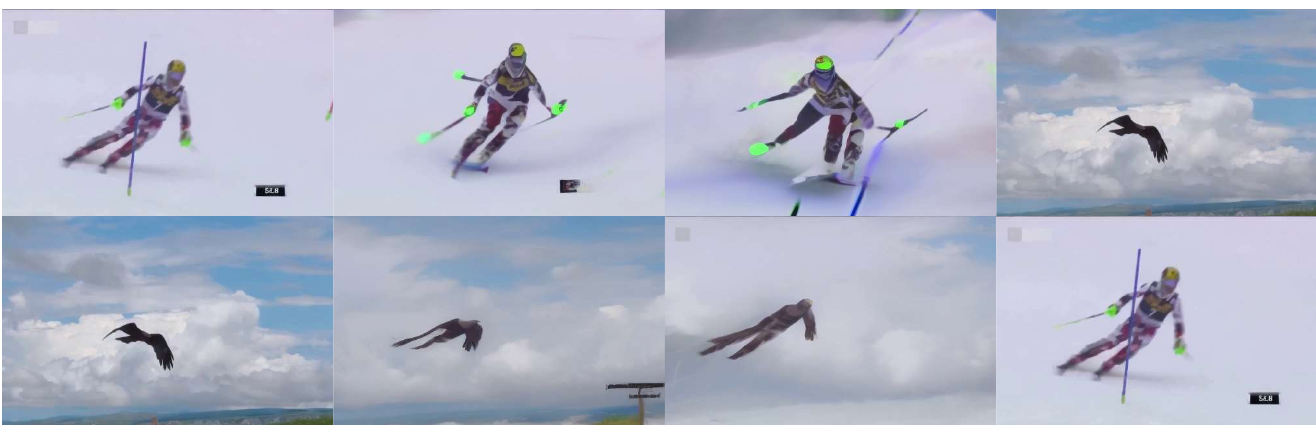}
    \caption{Samples of forward and reverse videos generated by DynamiCrafter, showing selected frames at 0, 5, 10, and 15. The top row shows the forward video, and the bottom row shows the reverse video.}
    \label{fig:reverse_samples}
\end{figure}

In practice, we observed significant quality differences in transition videos generated from two images, $x_0$ and $x_S$, depending on whether the sequence was processed in original or reverse order. As shown in Figure~\ref{fig:reverse_samples}, this issue is linked to conditional image leakage, where the transition video tends to resemble the initial image more closely. To address the shortcomings in the latter part of the video, particularly the deviations from the final frame and the lack of smoothness, we propose a Frequency-aware Bidirectional Fusion~(FBiF) approach to improve overall generation quality.

As illustrated in Figure~\ref{fig:framework}, we generate a forward transition video, then repeat the process by reversing the order of the input images and prompts fed into the video generation model. In the latent space, we apply $\mathcal{GPR}$ to map features into a unified Gaussian space, aligning the feature distributions of both the forward and reversed videos. Let $z_g = \mathcal{GPR}(z)$, where $z_g \in \mathbb{R}^{S \times N \times P}$; the reversed features are similarly defined as $z_g^{rev} \in \mathbb{R}^{S \times N \times P}$. Inspired by~\cite{pan2022fast}, we consider the average-pooled (avgpool) features to represent low-frequency components, while the max-pooled (maxpool) features capture high-frequency components. Since our goal is to emphasize the primary concepts in the video content rather than merely enhancing details, we employ a frequency-aware fusion method, combining the two feature sets along the temporal sequence as follows:
\begin{footnotesize}
\begin{equation} \label{eq:fusion}
    \begin{aligned}
    z_{fused}^s = &\ \lambda_s \cdot \text{avgpool}(z_g^s) 
    + (1-\lambda_s) \cdot \text{avgpool}(\text{reverse}(z_g^{rev})^s) \\
    &+ \lambda_{freq} \cdot \text{maxpool}(z_g^s) 
    + \lambda_{freq} \cdot \text{maxpool}(\text{reverse}(z_g^{rev})^s).
    \end{aligned}
\end{equation}
\end{footnotesize}

\noindent where $z_{fused}^s$ represents the fused feature at frame $s$. The weighting factor $\lambda_s$, which varies across frames from 0.9 to 0.1, controls the contributions of the low-frequency components. The high-frequency components are weighted by a fixed factor $\lambda_{freq}$. To ensure consistent video generation, we propose a Close-distance Frame Selection mechanism. Starting with the first frame of the forward video as a reference, we reverse the sequence of the reversed video. For each subsequent frame, we compare the perceptual loss~\cite{DBLP:conf/cvpr/ZhangIESW18} between the next frames of both videos relative to the previous frame. The frame with the smallest distance is selected as the next frame, and this process is repeated to generate the final video.

\section{Experiments}\label{sec:experiments_4}
\subsection{Experimental Setup}
We implemented our Transition Video Generation approach using the pre-trained DynamiCrafter model~\cite{xing2023dynamicrafter}. Most hyperparameters were kept at default settings, except for the FPS, which was set to 16 to enhance visual dynamics. The hyperparameter $\gamma$ was set to 0.9,  $\lambda_{freq}$  was 0.1. We used a DDIM~\cite{DBLP:conf/iclr/SongME21} schedule with 10 steps during denoising. All experiments ran on a single NVIDIA A800 GPU.

For quantitative evaluation, we tested the MorphBench~\cite{Zhang_2024_CVPR} and TC-Bench-I2V~\cite{DBLP:journals/corr/abs-2406-08656} datasets, generating 16 frames per video following established protocols~\cite{DBLP:conf/iclr/Chen0ZZMY0L0024, Zhang_2024_CVPR,DBLP:journals/corr/abs-2406-08656}. The video resolution was set to $512 \times 512$ for MorphBench and $320 \times 512$ for TC-Bench-I2V. For qualitative evaluation, we collected 10 image pairs from the Internet for comparison with other models and commercial products.

\subsection{Qualitative results}
\begin{figure}[htbp]
	\centering
	\includegraphics[width=\columnwidth]{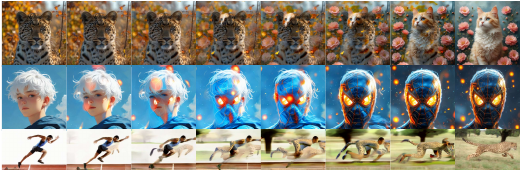}
	\caption{Visualization of Our Generated Samples.}
	\label{fig:Qualitative_results}
\end{figure}

To demonstrate the effectiveness of our algorithm, we have presented some of the results generated by our model in Figure~\ref{fig:Qualitative_results}. It can be seen that for different scenarios, our model exhibits strong video-transition capabilities, with the video maintaining dynamic elements while the content gradually transitions. Additionally, we compared our algorithm with several commercial products as shown, some of which were even unable to produce transition videos as shown in Figure~\ref{fig:failed_example}. For more visualization results, we recommend readers refer to the supplementary materials.

\subsection{Quantitative Results}

\textbf{Evaluation on TC-Bench-I2V:} We evaluated our model on the TC-Bench-I2V dataset using the Transition Completion Ratio (TCR, $\uparrow$) and TC-Score~($\uparrow$)~\cite{DBLP:journals/corr/abs-2406-08656}. These metrics utilize GPT-4 to provide assessments closely aligned with human perception. TCR measures the percentage of videos that successfully match the provided prompts, while TC-Score offers a comprehensive evaluation considering factors such as transition completion, object consistency, and handling of additional objects, reflecting user experiences more realistically.

Table~\ref{tab:TC-Bench-I2V} presents a comparison of our algorithm with three state-of-the-art methods across three video transition scenarios: \emph{Attribute Transition}~(Attribute), involving changes in color, shape, material, and texture; \emph{Object Relation Change}~(Object), focusing on object interactions like passing or striking; and \emph{Background Shifts}~(Background), such as transitions from day to night.

\begin{figure}[htbp]
	\centering
	\includegraphics[width=\columnwidth]{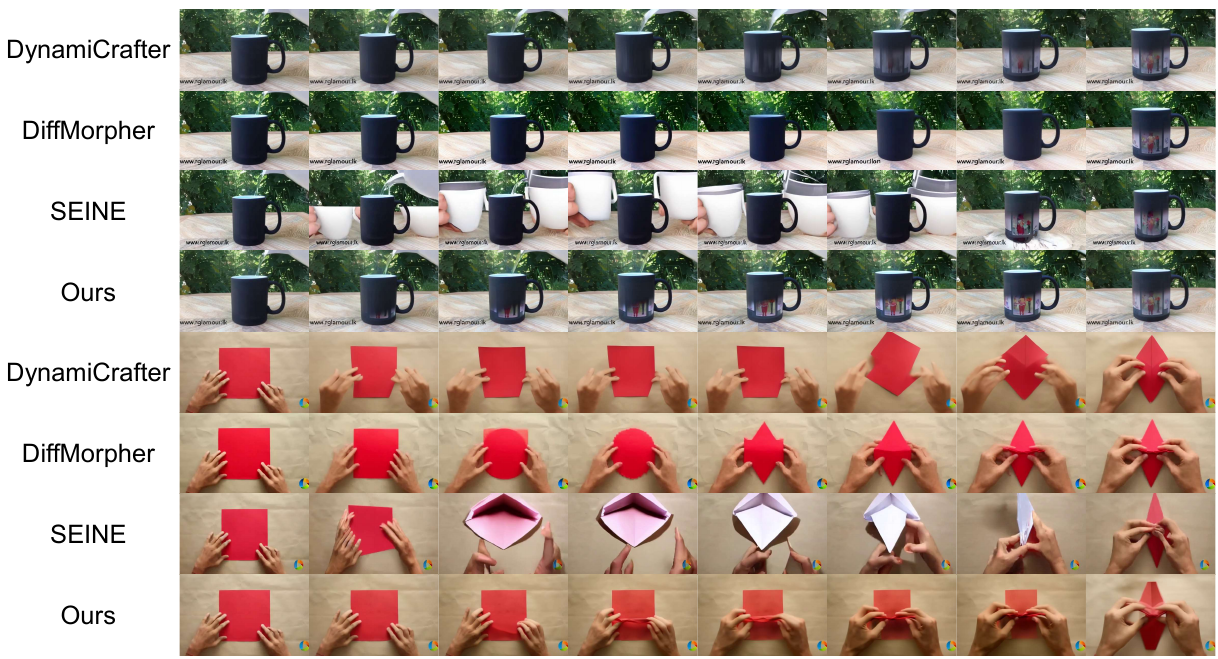}
	\caption{Comparison of Generated Sequences from TC-Bench Across Various Models.}
	\label{fig:tc-bench}
\end{figure}

Our results demonstrate consistent superior performance across most metrics and scenarios. Notably, our method ranks second in TCR for the background scenario, likely due to the lower dynamic demands, which favor DiffMorpher, a frame-by-frame generation-based approach. However, in dynamic scenarios like Object Relation Change, our algorithm significantly outperforms others, aligning more closely with user experience expectations. Figure~\ref{fig:tc-bench} illustrates comparative examples of generated results, with samples displayed every two frames.

\begin{table*}[htbp]
\centering
\caption{Quantitative evaluation on TC-Bench-I2V, We report TCR and TC-Score to assess the transition videos.}
\label{tab:TC-Bench-I2V}
\resizebox{1.4\columnwidth}{!}{%
\begin{tabular}{ccccccccc}
\hline
\multirow{2}{*}{Methods} & \multicolumn{2}{c}{Attribute}   & \multicolumn{2}{c}{Object}      & \multicolumn{2}{c}{Background}  & \multicolumn{2}{c}{Overall}     \\ \cline{2-9} 
& TCR$\uparrow$            & TC-Score~$\uparrow$       & TCR~$\uparrow$            & TC-Score~$\uparrow$       & TCR~$\uparrow$            & TC-Score~$\uparrow$       & TCR~$\uparrow$    & TC-Score~$\uparrow$       \\ \hline
DynamiCrafter~\shortcite{xing2023dynamicrafter}             & 16.55          & 0.745          & 13.91          & 0.707          & 25.56          & 0.795          & 16.89          & 0.738          \\
SEINE~\shortcite{DBLP:conf/iclr/Chen0ZZMY0L0024}                    & 17.86          & 0.720          & 10.48          & 0.654          & 7.96           & 0.742          & 13.57          & 0.698          \\
DiffMorpher~\shortcite{Zhang_2024_CVPR}                & \textbf{41.82}          & 0.844          & 19.57          & 0.765          & \textbf{50.00} & 0.819          & 34.45          & 0.810          \\ \hline
\textbf{Ours}            & \textbf{41.82} & \textbf{0.877} & \textbf{30.44} & \textbf{0.822} & 38.89          & \textbf{0.864} & \textbf{36.98} & \textbf{0.853} \\ \hline
\end{tabular}%
}
\end{table*}


\textbf{Evaluation on MorphBench:} Following previous work~\cite{Zhang_2024_CVPR}, we evaluated the fidelity and smoothness of video content using traditional metrics, specifically Frechet Inception Distance (FID, $\downarrow$)~\cite{DBLP:conf/nips/HeuselRUNH17} and Perceptual Path Length (PPL, $\downarrow$)~\cite{DBLP:conf/cvpr/KarrasLAHLA20}. The results on the MorphBench dataset, summarized in Table~\ref{tab:MorphBench}, show that while our approach does not outperform all existing methods, it remains competitive with current diffusion-based models~\cite{DBLP:conf/iclr/SongME21, DBLP:journals/corr/abs-2307-12560, Zhang_2024_CVPR, DBLP:conf/iclr/Chen0ZZMY0L0024, xing2023dynamicrafter}. Notably, even traditional techniques like Warp\&Blend surpass various generative algorithms in certain metrics, highlighting that superior quantitative scores do not necessarily translate to better visual effects in video transitions.

Our primary objective is to develop an algorithm that seamlessly transitions between the final frame of one video and the first frame of another, enhancing continuity and fluidity. Although dynamic videos, which are favored for their rich visual appeal~\cite{roggeveen2015impact, steuer1995defining, coyle2001effects}, often introduce greater frame variation that can reduce quantitative performance, the enhanced user experience remains paramount.

\begin{figure*}[htbp]
	\centering
	\includegraphics[width=2\columnwidth]{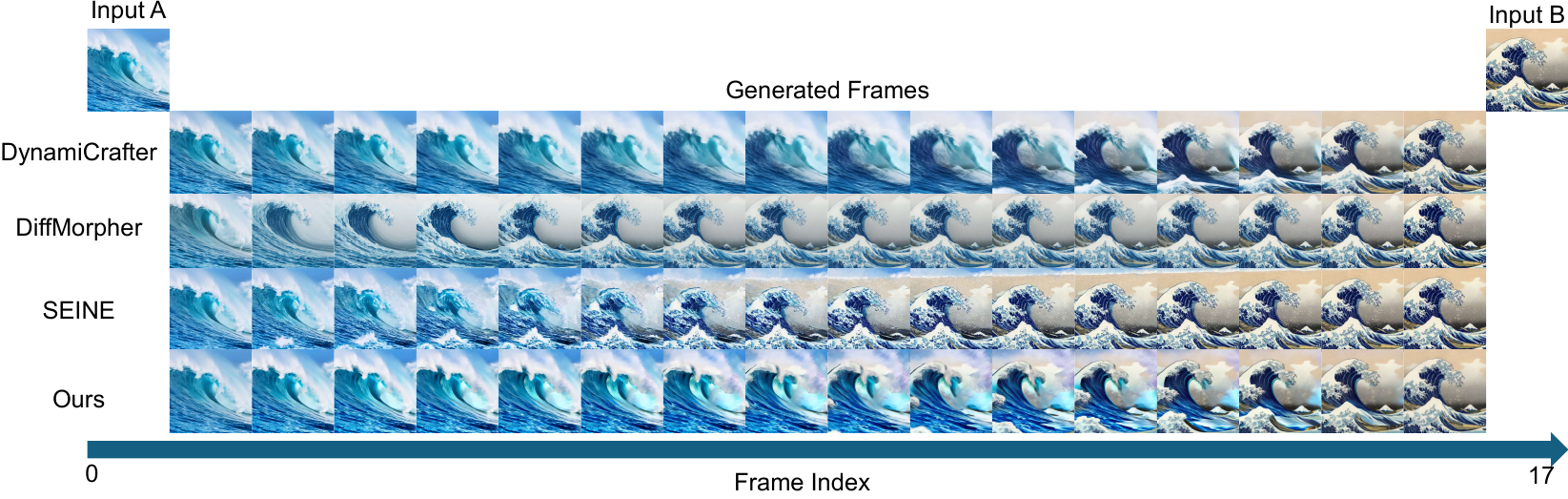}
	\caption{Visualization of Wave Sequence from the MorphBench dataset compared with other methods. Our method achieves gradual changes while preserving motion.}
	\label{fig:morph_wave}
\end{figure*}

\begin{figure}[htbp]
	\centering
	\includegraphics[width=\columnwidth]{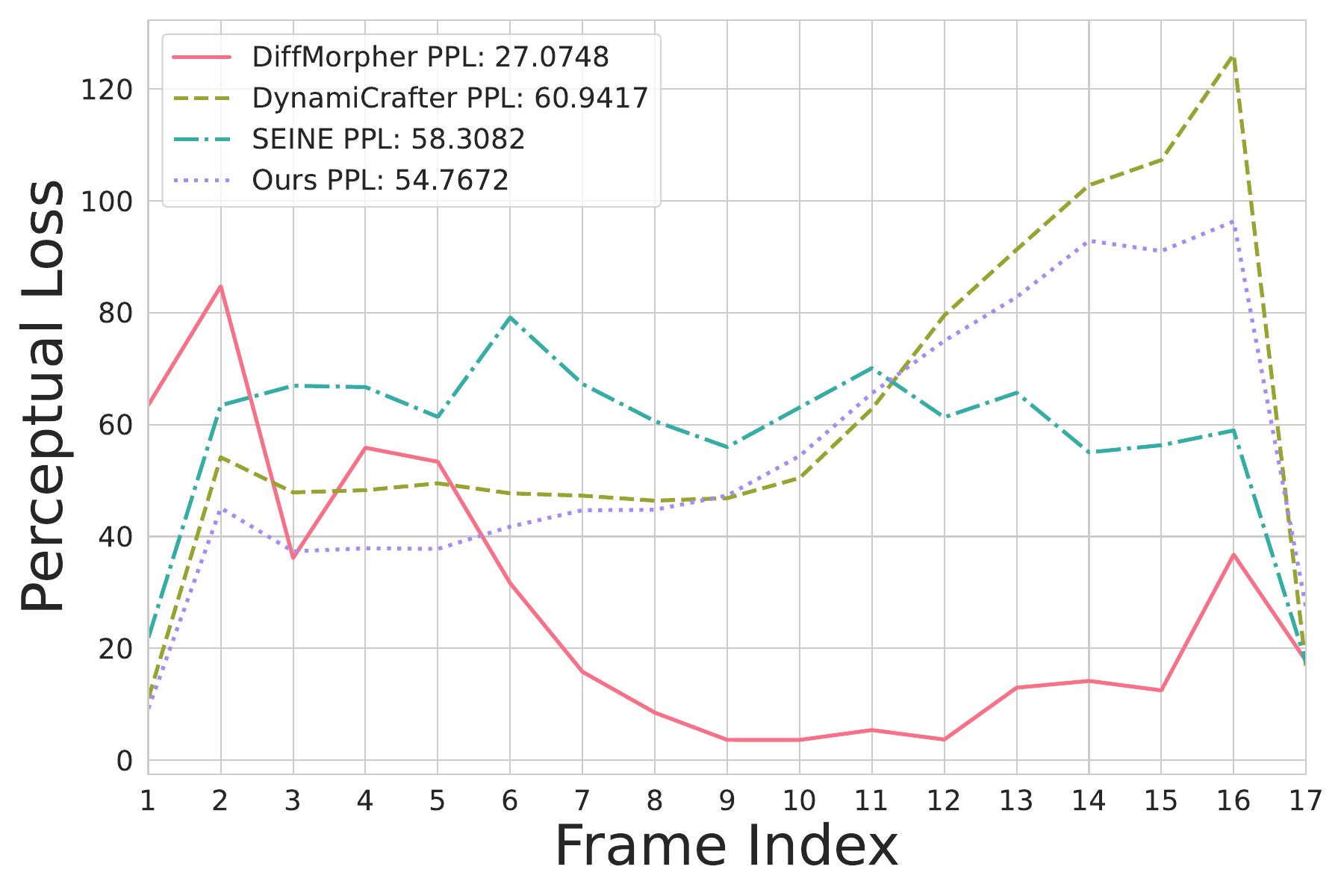}
	\caption{Perceptual Losses between Generated Frames of Wave Sequence in MorphBench}
	\label{fig:morph_wave_ppl}
\end{figure}

Figure~\ref{fig:morph_wave} illustrates content generated from a pair of images in the MorphBench dataset, using identical prompts across different methods. Our method and DiffMorpher employ separate prompts for each image, while SEINE and DynamiCrafter merge the descriptions into a single prompt, transitioning from input A to input B without introducing additional content. Figure~\ref{fig:morph_wave_ppl} presents the perceptual loss~\cite{DBLP:conf/cvpr/ZhangIESW18} between consecutive frames and final PPL, starting with input A at frame index 0 and ending with input B at frame index 17, covering both input images and generated frames. DiffMorpher achieves the lowest PPL, indicating superior quantitative performance. This is primarily because only a small portion of the generated frames retain perceptual similarity to input A, with the rest transitioning gradually toward input B in a low-dynamic manner, resulting in final frames that closely resemble the target. Although perceptual loss is relatively high during the initial dynamic phase, it remains low thereafter, enhancing DiffMorpher's quantitative metrics. However, this approach results in a near-static user experience, making it difficult to effectively engage the audience.

In contrast, other methods, including ours, prioritize transforming image characteristics while maintaining high dynamic variability, leading to fluctuating PPL values throughout the sequence and impacting the final quantitative results. While videos with high dynamic variability often appeal more to users, SEINE and DynamiCrafter demonstrate more abrupt transitions. SEINE fully transitions to input B midway through the sequence with minimal change afterward, while DynamiCrafter adopts the style of input B only in the final quarter, lacking the smoothness needed for a seamless transition. In contrast, our method excels at gradually blending specific regions into the style of input B while preserving natural motion, achieving both high dynamic performance and smooth, visually coherent transitions throughout the entire sequence, underscoring the superiority of our approach.

Another factor contributing to the significant perceptual loss between frames in highly dynamic content is the limitation of existing open-source video generation models, which struggle to produce longer video sequences. Typically, these models generate around 16 frames, and content beyond this length often becomes unstable. As our method is training-free, it currently cannot be extended to generate videos with more frames. Ensuring a high dynamic range under this frame limitation inevitably impacts the quality and smoothness of the video. However, this impact is acceptable under current circumstances. If more advanced open-source models for generating longer videos become available, our method should achieve further performance improvements.

\begin{table}[thbp]
\centering
\caption{Quantitative evaluation on MorphBench, We report FID and PPL to assess the fidelity and smoothness of the transition videos, respectively.  \textbf{Bold} indicates the best result, \underline{underline} indicates the second best result.
}
\label{tab:MorphBench}
\resizebox{\columnwidth}{!}{%
\begin{tabular}{ccccccc}
\hline
\multirow{2}{*}{Method} & \multicolumn{2}{c}{Metamorphosis} & \multicolumn{2}{c}{Animation} & \multicolumn{2}{c}{Overall} \\ \cline{2-7} 
                        & FID~$\downarrow$             & PPL~$\downarrow$             & FID~$\downarrow$            & PPL~$\downarrow$          & FID~$\downarrow$          & PPL~$\downarrow$          \\ \hline
Warp\&Blend~\shortcite{DBLP:journals/vc/Wolberg98}             & \underline{79.63}           & \textbf{15.97}           & 56.86          & \textbf{9.58}         & 67.57        & \textbf{14.27}        \\
DGP~\shortcite{DBLP:journals/pami/PanZDLLL22}                     & 150.29          & 29.65           & 194.65         & 27.50        & 138.20       & 29.08        \\
StyicGAN-XL~\shortcite{DBLP:conf/siggraph/SauerS022}             & 122.42          & 41.94           & 133.73         & 33.43        & 112.63       & 39.67        \\
DDIM~\shortcite{DBLP:conf/iclr/SongME21}                    & 95.44           & 27.38          & 174.31         & 18.70        & 101.68       & 25.38        \\
DiffInterp~\shortcite{DBLP:journals/corr/abs-2307-12560}              & 169.07          & 108.51          & 148.95         & 96.12        & 146.66       & 105.23       \\
DynamiCrafter~\shortcite{xing2023dynamicrafter}           & 87.32           & 42.09           & 43.31          & 11.16        & 69.13        & 33.84        \\
DiffMorpher~\shortcite{Zhang_2024_CVPR}             & \textbf{70.49}           & \underline{18.19}           & \underline{43.15}          & \textbf{5.14}         & \textbf{54.69}        & \underline{21.10}        \\
SEINE~\shortcite{DBLP:conf/iclr/Chen0ZZMY0L0024}                   & 82.03           & 47.72           & 48.25          & 16.26        & 67.60        & 39.33        \\  \hline
Ours                    & 86.92           & 35.18           & \textbf{42.99}          & 12.46        & \underline{64.05}        & 29.08        \\ \hline
\end{tabular}%
}
\end{table}

\textbf{Human Evaluation:} To assess the real-world user experience of our algorithm, we conducted a Human Evaluation study using 22 curated videos: 9 from MorphBench, 9 from TC-Bench, and 4 from our own dataset. 15 volunteers participated, yielding a total of 990 votes. Participants compared pairs of videos and voted on which appeared more natural and smooth across three scenarios: \emph{Ours vs. DiffMorpher}, \emph{Ours vs. SEINE}, and \emph{Ours vs. DynamiCrafter}. The results, summarized in Table~\ref{tab:human-evaluation}, indicate a strong preference for our algorithm, highlighting its effectiveness in generating smoother, more natural transitions.

\begin{table}[htbp]
\centering
\caption{Human Preference on transition video quality.}
\label{tab:human-evaluation}
\resizebox{\columnwidth}{!}{%
\begin{tabular}{cccc}
\hline
Metrics    & Ours \textgreater DiffMorpher & Ours \textgreater SEINE & Ours \textgreater DynamiCrafter \\ \hline
Perference & 71.82\%                              & 75.45\%                        &  75.45\%                   \\ \hline
\end{tabular}%
}
\end{table}

\section{Ablation Study}

\begin{figure}[htbp]
	\centering
	\includegraphics[width=0.9\columnwidth]{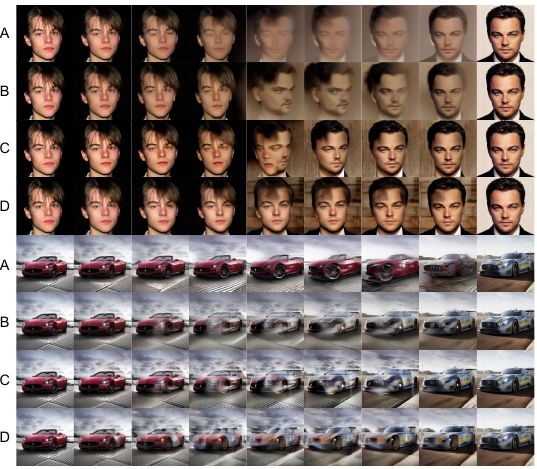}
	\caption{Visualization of Ablation Study.}
	\label{fig:abl_study}
\end{figure}

To validate the effectiveness of our proposed module, we conducted ablation studies on the MorphBench dataset. The results, presented in Table~\ref{tab:Ablation-evaluation} and Figure~\ref{fig:abl_study}, highlight the impact of each component on overall performance. Specifically, Method A represents the baseline model, DynamiCrafter; Method B incorporates interpolation-based conditional controls; Method C integrates $\mathcal{GPR}$; and Method D corresponds to our final model. As shown in Table~\ref{tab:Ablation-evaluation}, our approach outperforms others across key quantitative metrics. Although these metrics may not always directly correlate with final visual quality, they strongly suggest that our module significantly enhances the baseline model's performance, especially in a training-free setting.

The qualitative results in Figure~\ref{fig:abl_study} further illustrate the effectiveness of our approach. In some videos, the baseline model fails to generate appropriate transitions, leading to abrupt changes or blurred content. Incorporating interpolation-based conditional controls improves transition content, though some issues with blurred frames remain. The integration of $\mathcal{GPR}$ substantially improves visual quality, although it occasionally introduces artifacts. Finally, with the addition of the FBiF module, most videos exhibit smooth and visually coherent transitions, effectively addressing these challenges.

\begin{table}[htbp]
\centering
\caption{Ablation studies of our method.}
\label{tab:Ablation-evaluation}
\resizebox{0.8\columnwidth}{!}{%
\begin{tabular}{ccccccc}
\hline
\multirow{2}{*}{Method}                & \multicolumn{2}{c}{Metamorphosis} & \multicolumn{2}{c}{Animation} & \multicolumn{2}{c}{Overall} \\ \cline{2-7} 
                                       & FID~$\downarrow$             & PPL~$\downarrow$             & FID~$\downarrow$           & PPL~$\downarrow$           & FID~$\downarrow$          & PPL~$\downarrow$          \\ \hline
A                          & 87.32           & 42.09           & 43.31         & \textbf{11.16 }        & 69.13        & 33.84        \\
B                    & 89.50           & 35.68           & 35.84         & 11.27         & 70.30        & 29.17        \\
C             & 90.51           & 36.25           & \textbf{42.05}         & 12.45         & 72.13        & 29.90        \\
D & \textbf{86.92}           & \textbf{35.18}           & 42.99         & 12.46         & \textbf{64.05}        & \textbf{29.08}        \\ \hline
\end{tabular}%
}
\end{table}

\section{Conclusion}
\label{sec:Conclusion}
We introduced a novel training-free transition video generation approach that combines interpolation-based conditional controls, $\mathcal{GPR}$, and FBiF. Our method consistently outperforms existing models in both quantitative metrics and visual quality, especially in dynamic scenarios. Ablation studies confirm that each component enhances the overall performance, leading to smoother and more coherent transitions. This approach shows significant potential for improving video generation without additional training, particularly in high-dynamic transitions. In the future, we plan to explore the use of entire video segments, rather than just two frames, to achieve even more seamless and effective transitions.


%

\bibliography{aaai25}
\appendix

\section{Appendix A: Implementation Details}
In all of our experiments, we utilized the publicly available DynamiCrafter~\cite{xing2023dynamicrafter} (interpolation variation@$320\times512$ resolution). The experiments were conducted without any model training, leveraging a PyTorch-based implementation on an Ubuntu 22.04 system. Video generation of 16 frames at a resolution of 320$\times$512 using an A800 GPU required approximately 55 seconds, with a memory usage of about 15GB, demonstrating the system's relatively low resource consumption.

We employed the MorphBench~\cite{Zhang_2024_CVPR} and TC-Bench-I2V~\cite{DBLP:journals/corr/abs-2406-08656} datasets datasets for our evaluations, which . Due to the high cost associated with the TC-Bench-I2V dataset, which necessitates invoking the GPT-4 API, our evaluation was limited to comparative experiments, and no ablation studies were performed. However, the prompts used were sourced from the official dataset, ensuring fairness in the comparisons.

For the MorphBench dataset, we conducted a comprehensive evaluation, including ablation studies. The prompts were primarily simple descriptions of image content, with detailed prompts provided in the supplementary material. Due to the interdependencies between our modules, the FBiF module was not ablated individually, as the absence of $\mathcal{GPR}$ for unified mapping of the feature space would result in feature summation that could severely disrupt the generative model, rendering video content generation infeasible.

\section{Appendix B: Limitations}
\label{sec:Limitations}
Our model is a training-free approach, which allows us to enhance the transition video generation performance of pre-trained models while also being subject to the inherent limitations of the base model's generation capabilities. When employing a higher number of DDIM steps, such as 50 steps, existing image-to-video generation models are prone to a phenomenon known as conditional image leakage~\cite{zhao2024identifying}. This issue leads to a reduction in the dynamic content of the generated videos as the number of steps increases. Consequently, our method may occasionally struggle to produce a proper transition content under these conditions. However, our success rate still exceeds that of the original model.

\begin{figure*}[htbp]
	\centering
	\includegraphics[width=2\columnwidth]{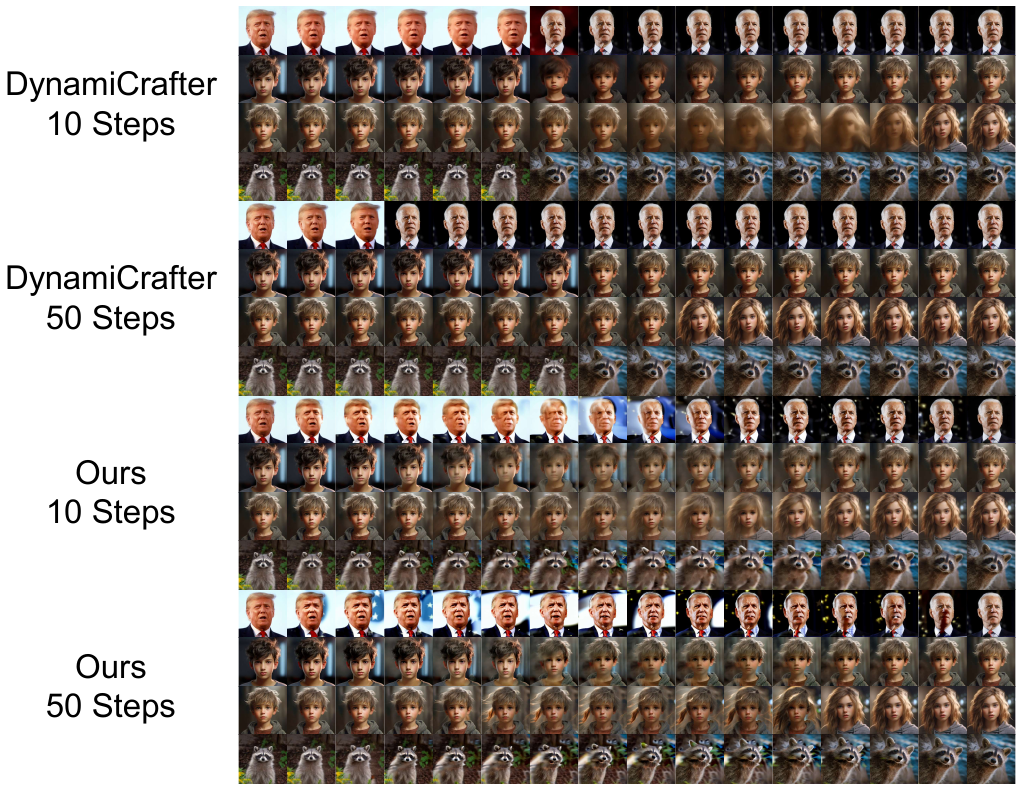}
	\caption{Visualization of Different Steps.}
	\label{fig:Limitations}
\end{figure*}

When using fewer steps, such as 10 steps, our method consistently outperforms the original model in terms of transition effects, as demonstrated in our experiments. Although there is a slight reduction in per-frame quality compared to the 50-step scenario, this difference does not significantly affect the user experience. As shown in Figure~\ref{fig:Limitations}, a comparison between some frames generated with 10 steps and those with 50 steps reveals that this quality difference is negligible in terms of perceptual impact on the user.

\section{Appendix C: User Study}
\label{sec:User_Study}

\begin{figure}[htbp]
	\centering
	\includegraphics[width=\columnwidth]{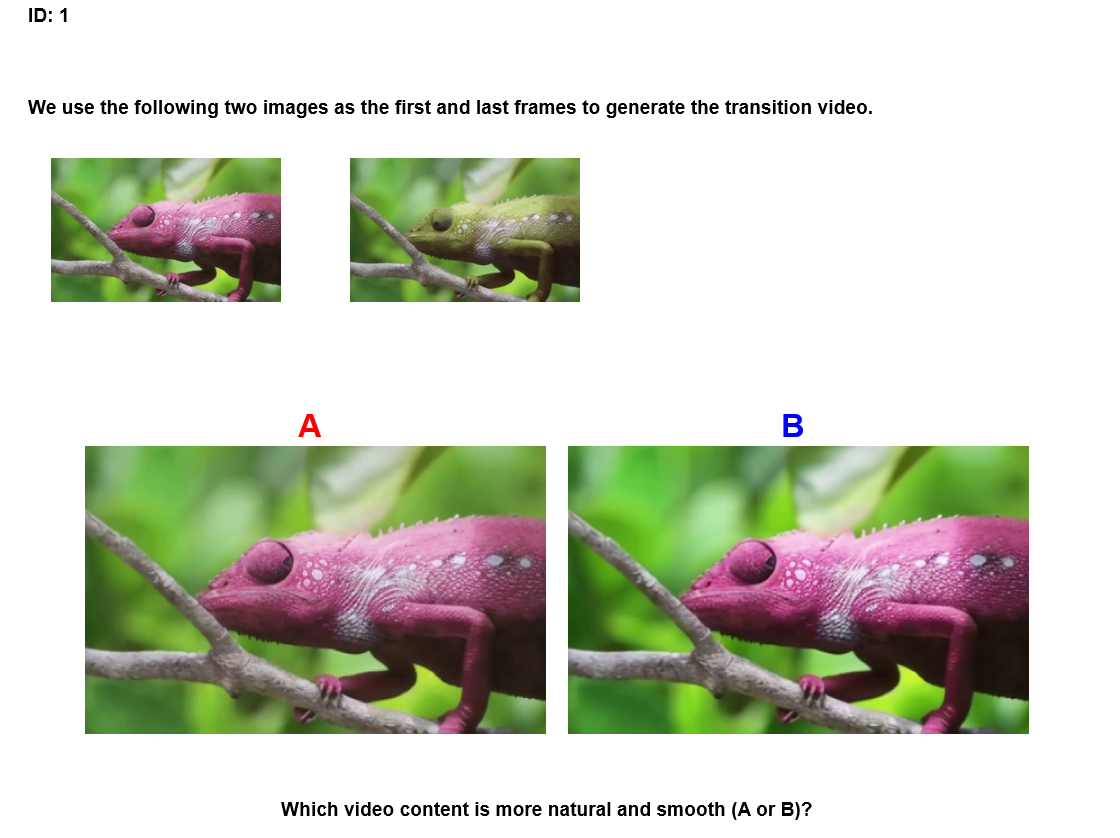}
	\caption{Example of the Questionnaire Content}
	\label{fig:user_questionnaire}
\end{figure}

\begin{figure}[htbp]
	\centering
	\includegraphics[width=\columnwidth]{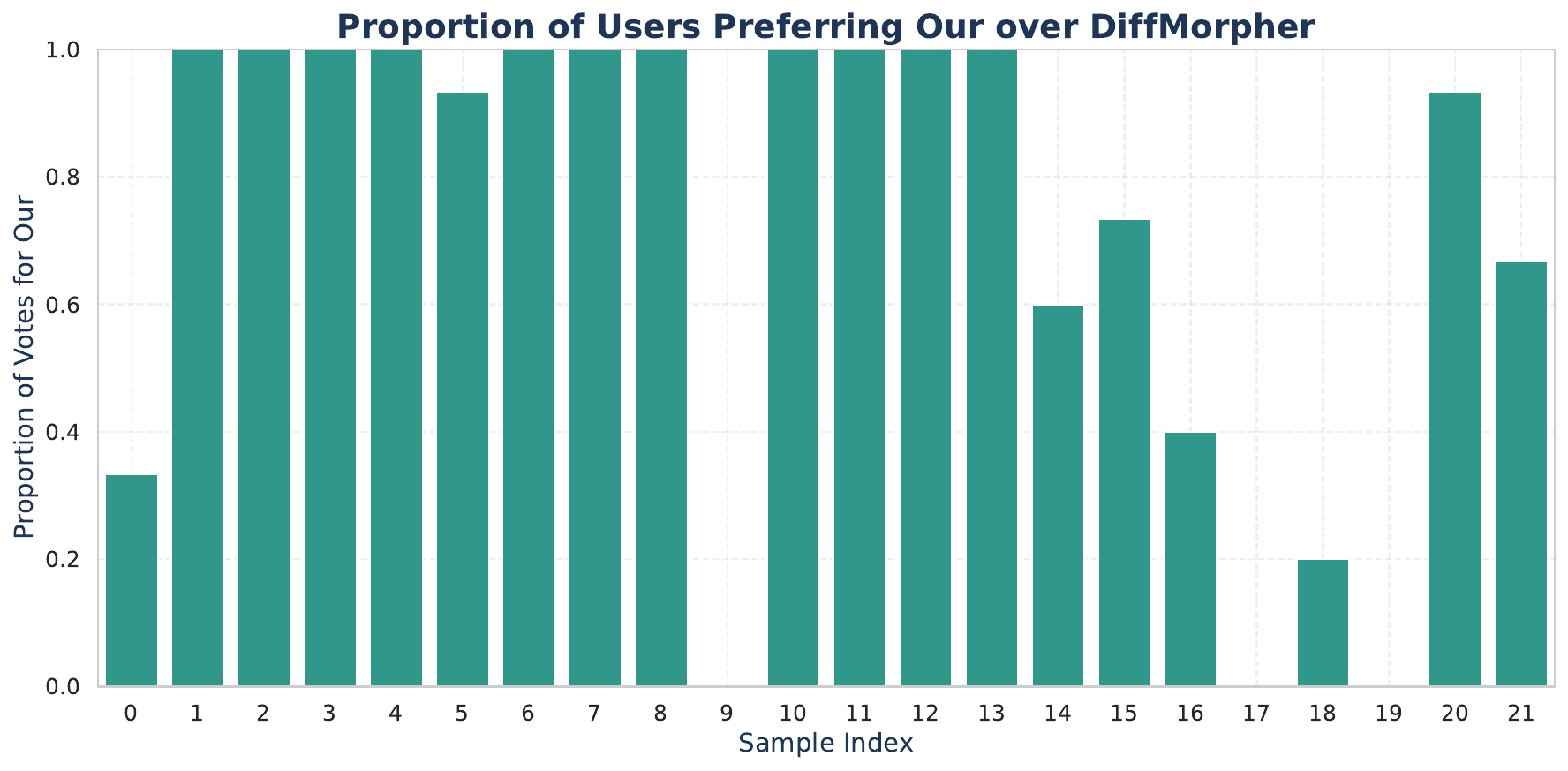}
	\caption{Ours \textit{vs.} DiffMorpher}
	\label{fig:user_DiffMorpher}
\end{figure}

\begin{figure}[htbp]
	\centering
        \includegraphics[width=\columnwidth]{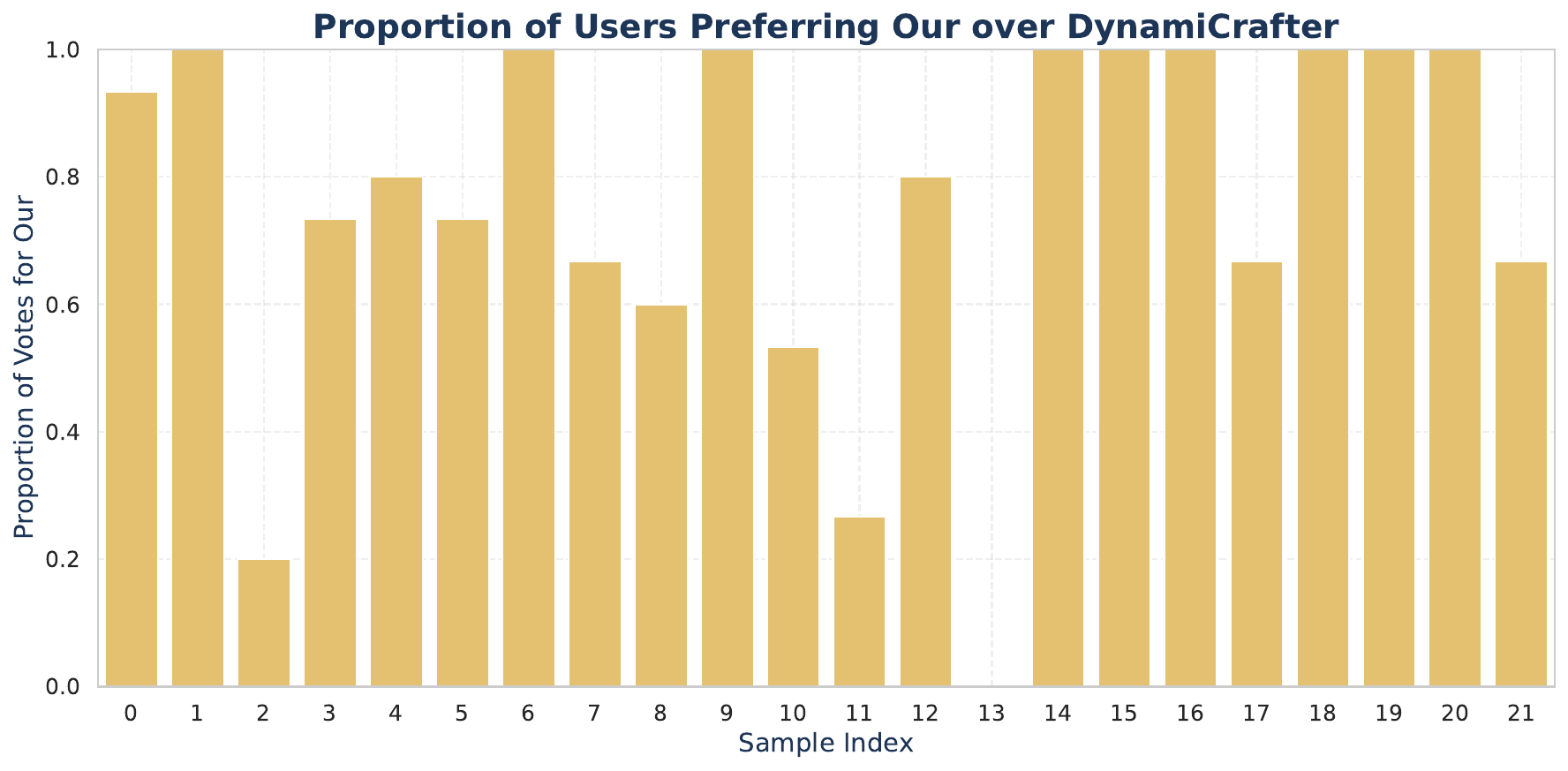}
        \caption{Ours \textit{vs.} DynamiCrafter}
	\label{fig:user_DynamiCrafter}
\end{figure}

\begin{figure}[htbp]
	\centering
        \includegraphics[width=\columnwidth]{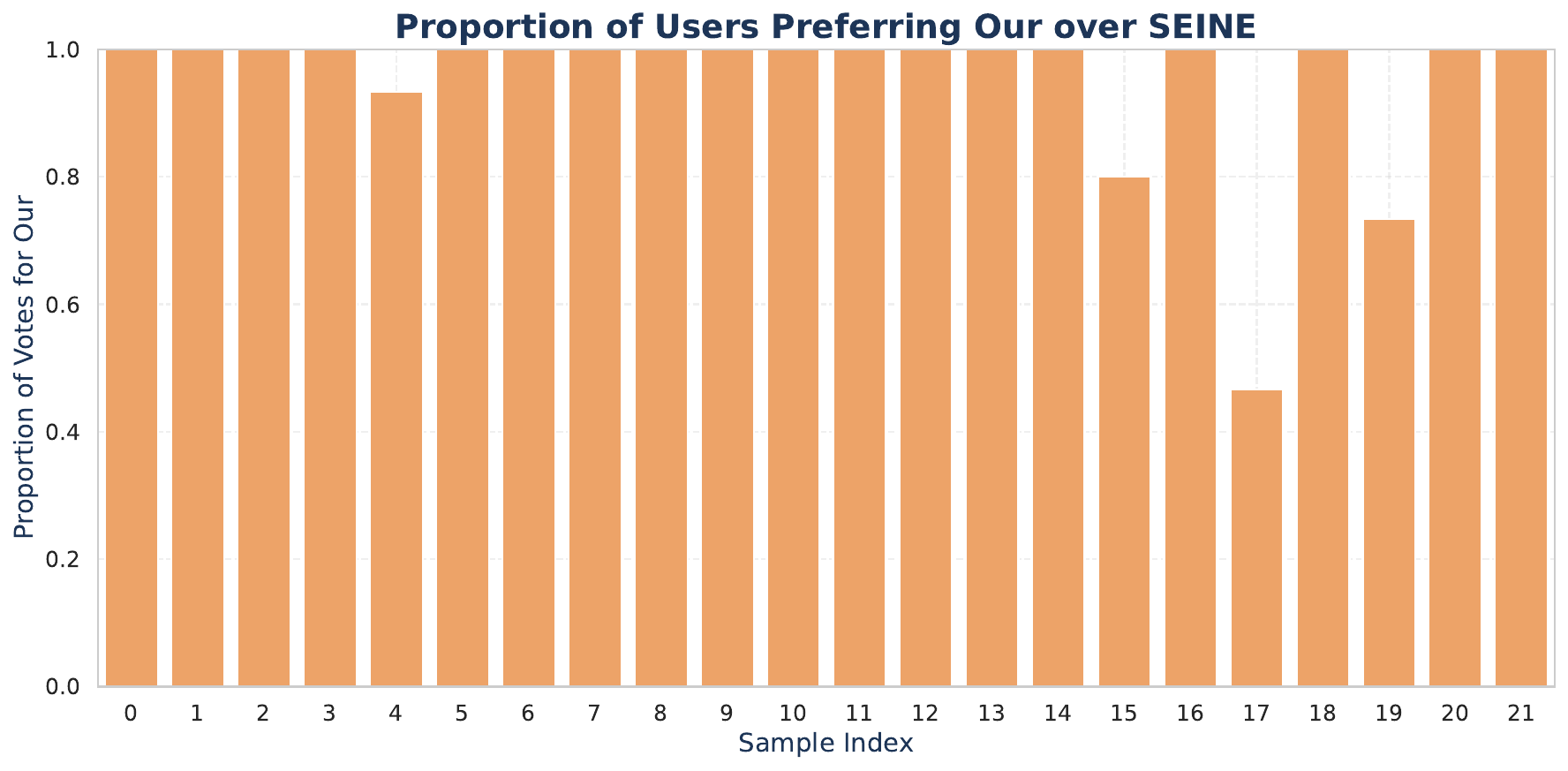}
        \caption{Ours \textit{vs.} SEINE}
	\label{fig:user_SEINE}
\end{figure}

Considering that quantitative metrics are insufficient to evaluate the appeal of transitions to users, which is more critical in real-world scenarios, we conducted a survey involving 15 volunteers. The survey included 22 randomly selected videos from our dataset, evaluated through three rounds of pairwise comparisons between our method and SEINE~\cite{DBLP:conf/iclr/Chen0ZZMY0L0024}, DiffMorpher~\cite{Zhang_2024_CVPR}, and DynamiCrafter~\cite{xing2023dynamicrafter}. Volunteers assessed the quality of the transition videos, with examples of the questionnaire content shown in Figure~\ref{fig:user_questionnaire}. The results, illustrating the preference for our method in each sample, are shown in Figure~\ref{fig:user_DiffMorpher},  Figure~\ref{fig:user_DynamiCrafter}and Figure~\ref{fig:user_SEINE},.

\begin{figure*}[htbp]
	\centering
	\includegraphics[width=2\columnwidth]{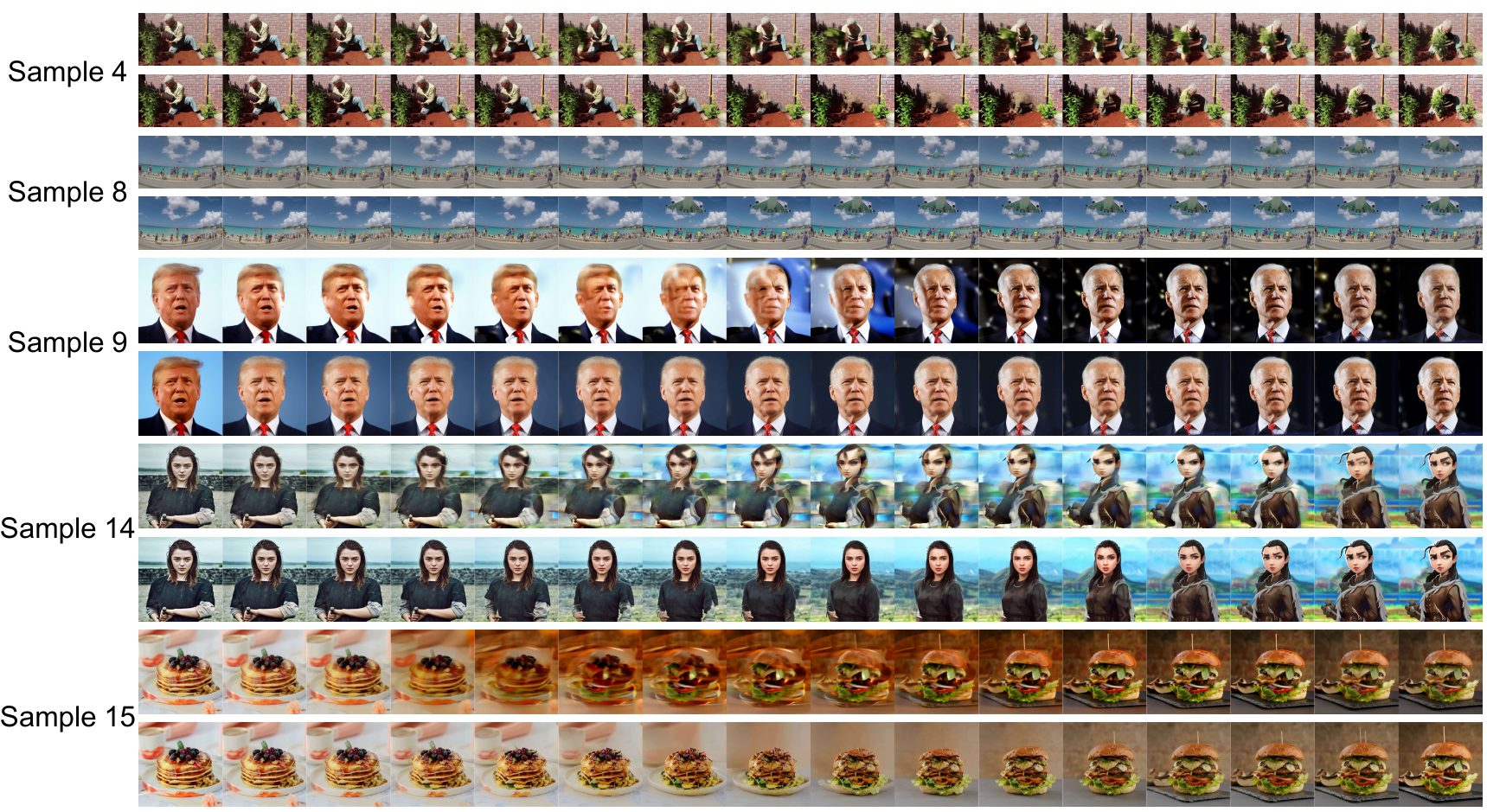}
	\caption{Comparison samples from our user study. The top row of each group shows results from our algorithm, while the bottom row shows results from DiffMorpher.}
	\label{fig:user_DiffMorpher2}
\end{figure*}

\begin{figure*}[htbp]
	\centering
	\includegraphics[width=2\columnwidth]{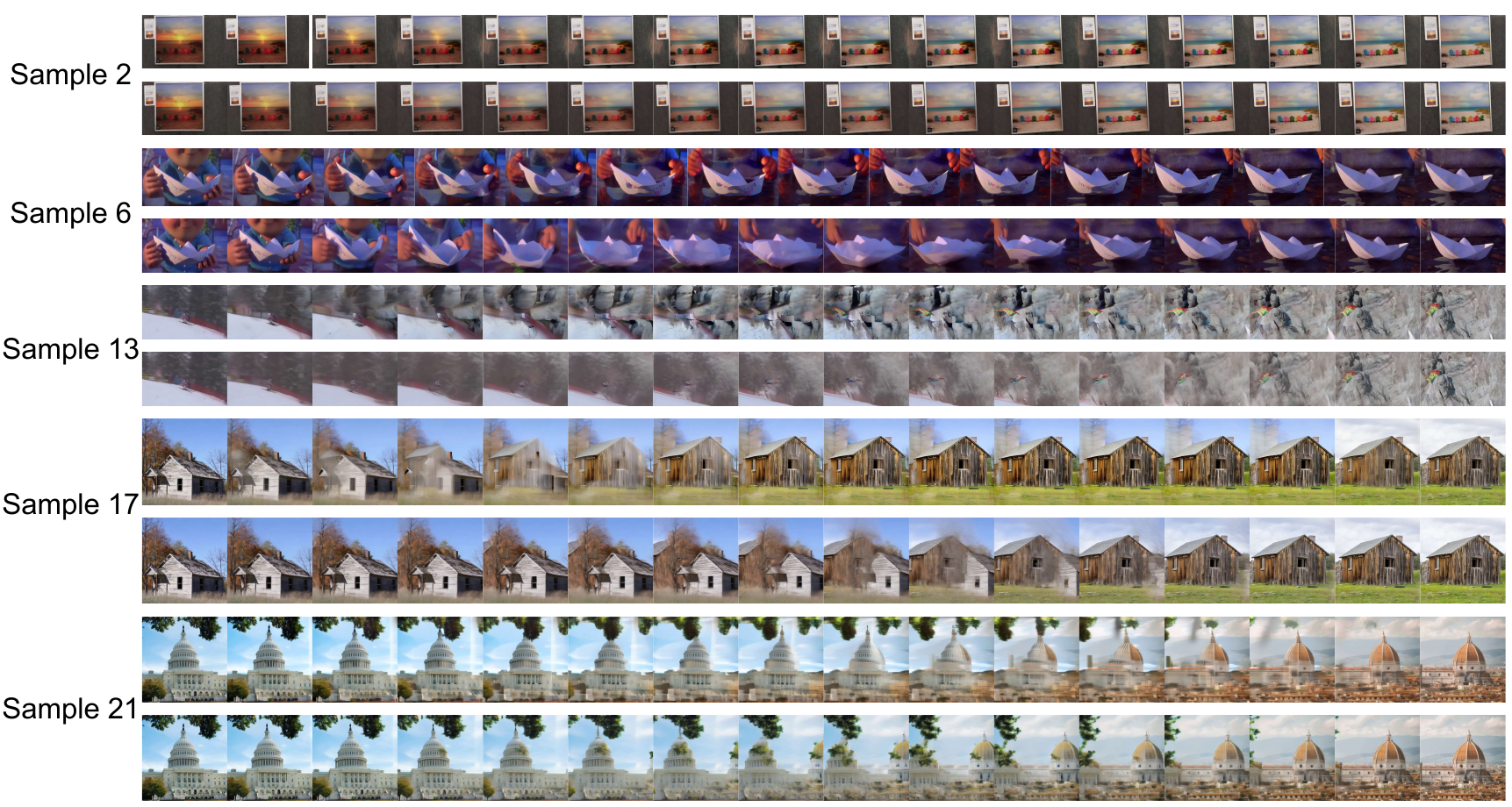}
	\caption{Comparison samples from our user study. The top row of each group shows results from our algorithm, while the bottom row shows results from DynamiCrafter.}
	\label{fig:user_DynamiCrafter2}
\end{figure*}

\begin{figure*}[htbp]
	\centering
	\includegraphics[width=2\columnwidth]{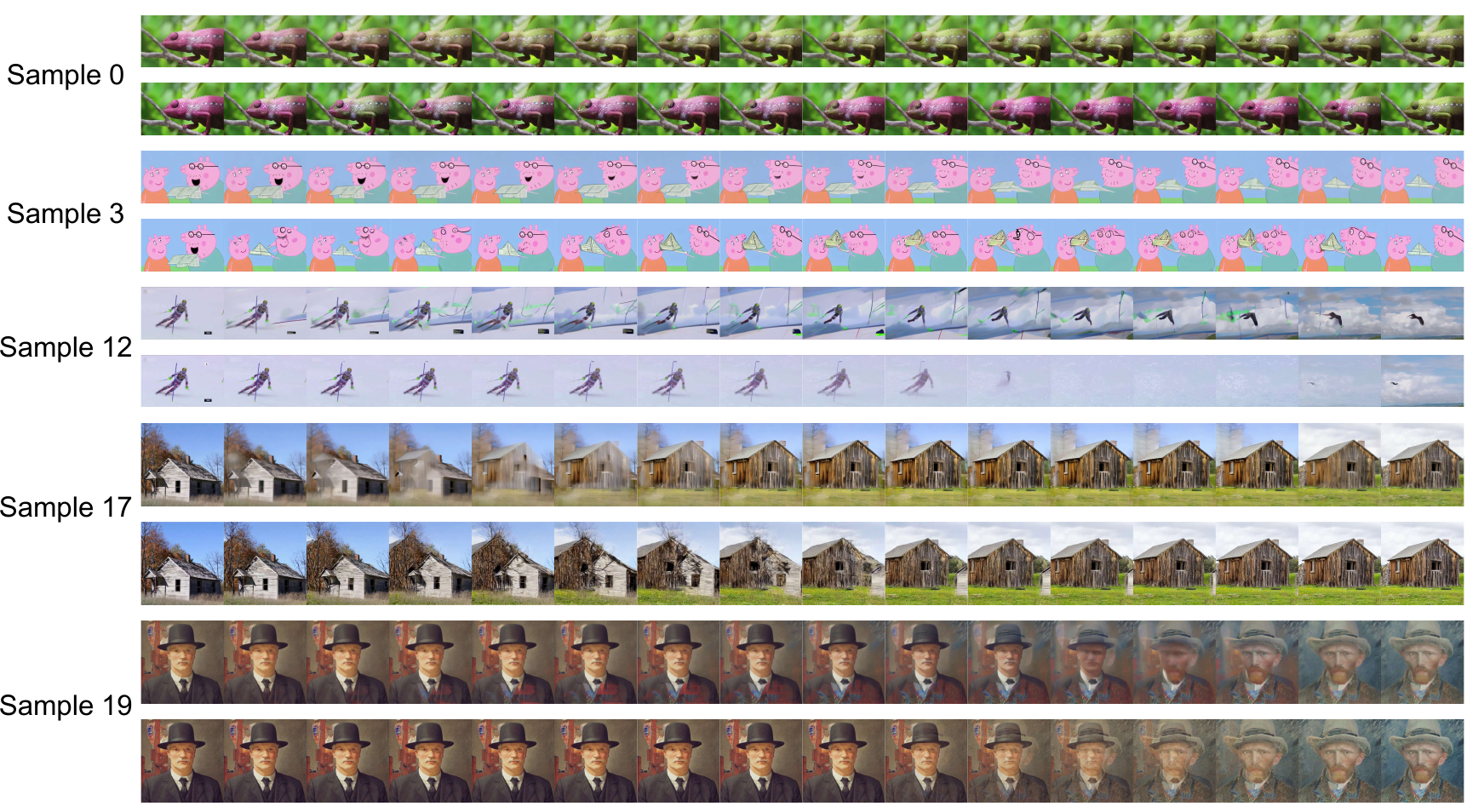}
	\caption{Comparison samples from our user study. The top row of each group shows results from our algorithm, while the bottom row shows results from SEINE.}
	\label{fig:user_SEINE2}
\end{figure*}

The results of the user study, combined with the findings in the main text, demonstrate that our method outperforms the other three algorithms in most cases. Among the compared methods, DiffMorpher exhibited slightly higher quality relative to SEINE and DynamiCrafter, aligning with the TC-Bench experimental results discussed earlier in the paper. We also provide examples from a subset of user evaluations, including cases with high user preference, cases with moderate disagreement (i.e., preference scores between 0.4 and 0.6), and cases with relatively lower quality samples, as illustrated in Figures~\ref{fig:user_DiffMorpher2}, \ref{fig:user_DynamiCrafter2}, and \ref{fig:user_SEINE2}. It is evident that our approach maintains effective video transitions regardless of user preference, demonstrating consistent dynamics. In contrast, other methods exhibit unstable transition effects, which are noticeably unacceptable in certain scenarios.


\section{Appendix D: Qualitative results}
\label{sec:Qualitative}

\begin{figure*}[htbp]
    \centering
    \includegraphics[width=2\columnwidth]{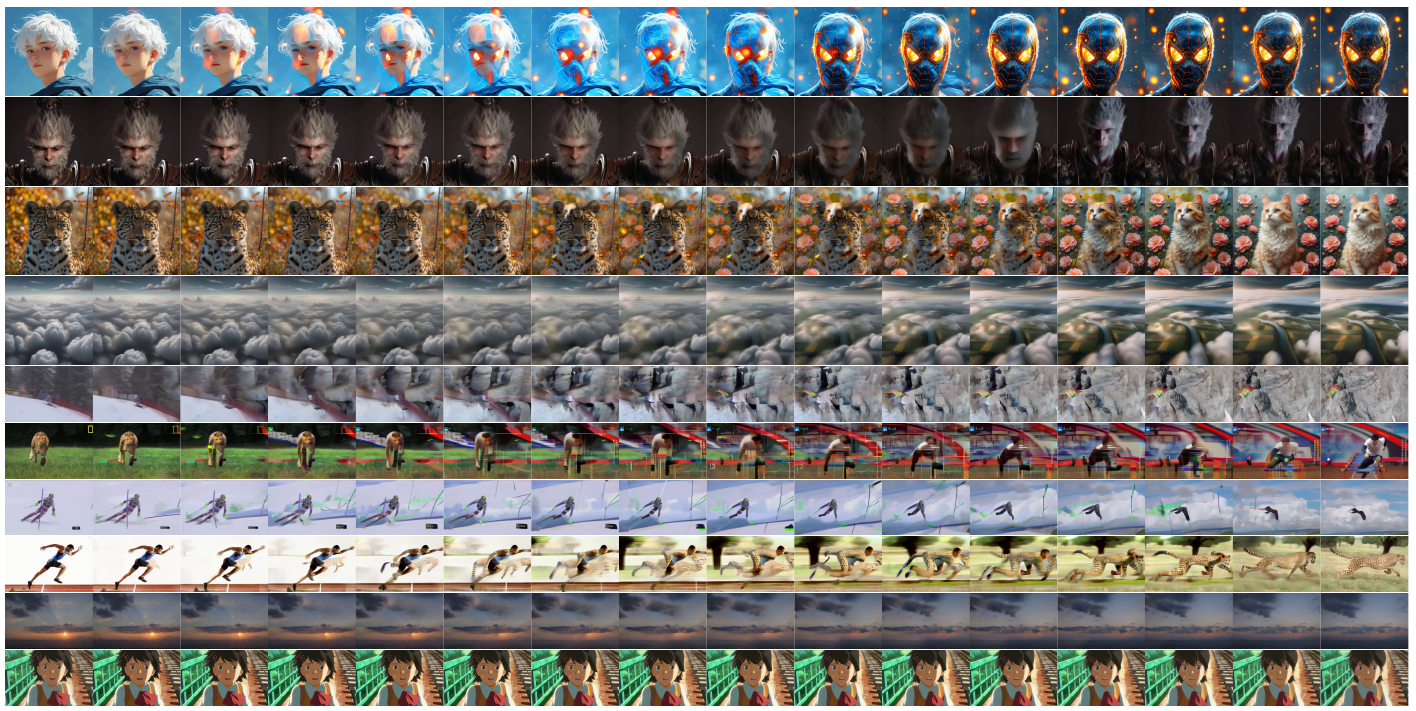}
    \caption{Generated Samples from Our Method}
    \label{fig:ours}
\end{figure*}

\begin{figure*}[htbp]
    \centering
    \includegraphics[width=2\columnwidth]{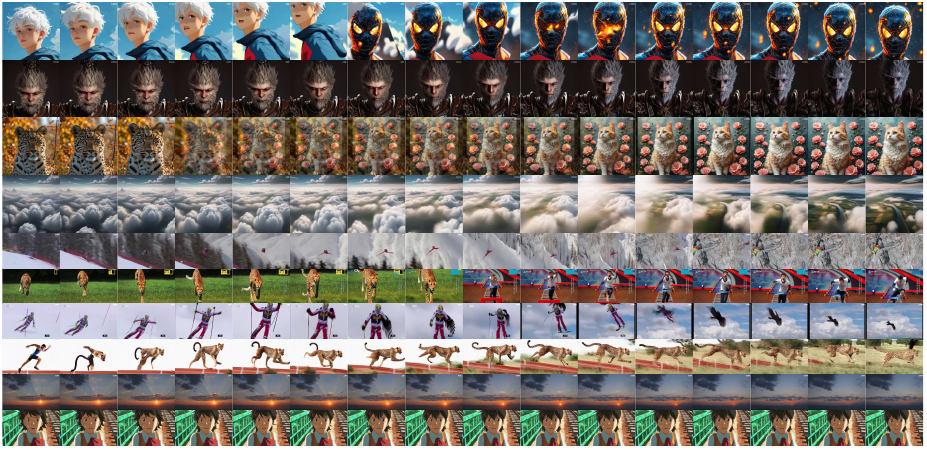}
    \caption{Generated Samples from LUMA AI}
    \label{fig:luma}
\end{figure*}

\begin{figure*}[htbp]
    \centering
    \includegraphics[width=2\columnwidth]{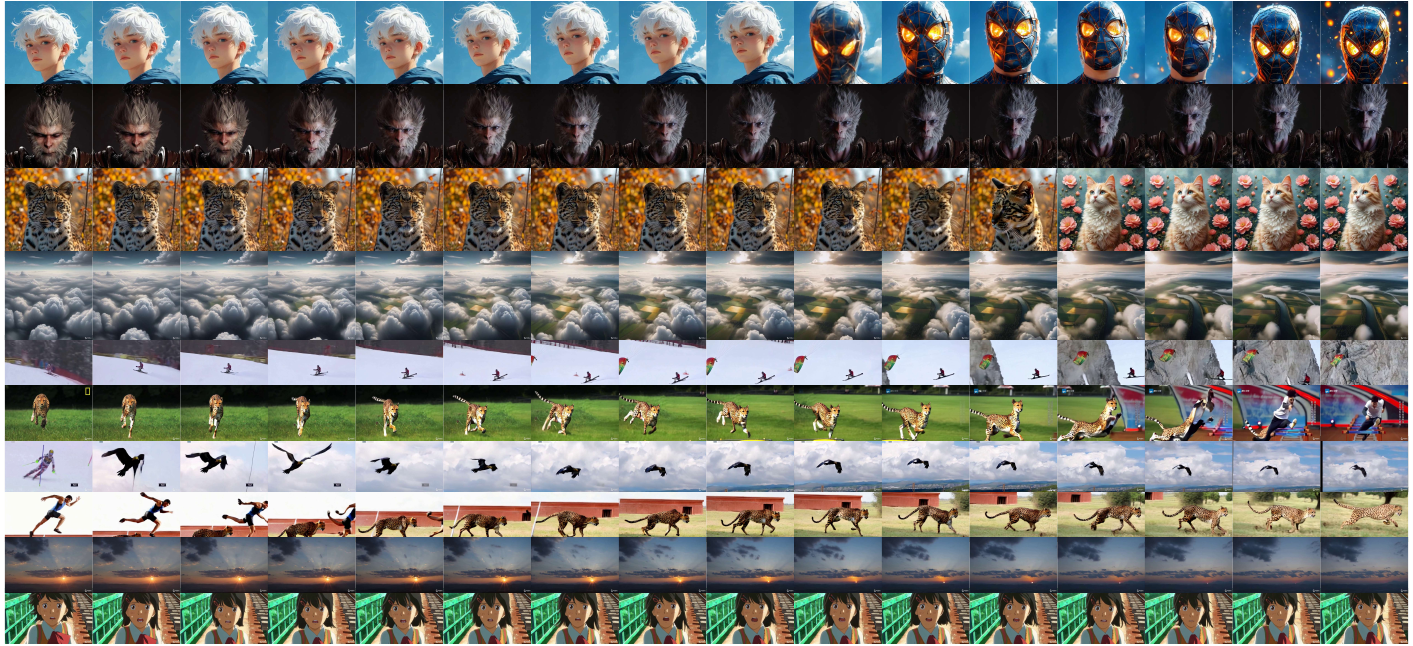}
    \caption{Generated Samples from KLing AI}
    \label{fig:kling}
\end{figure*}

\begin{figure*}[htbp]
    \centering
    \includegraphics[width=2\columnwidth]{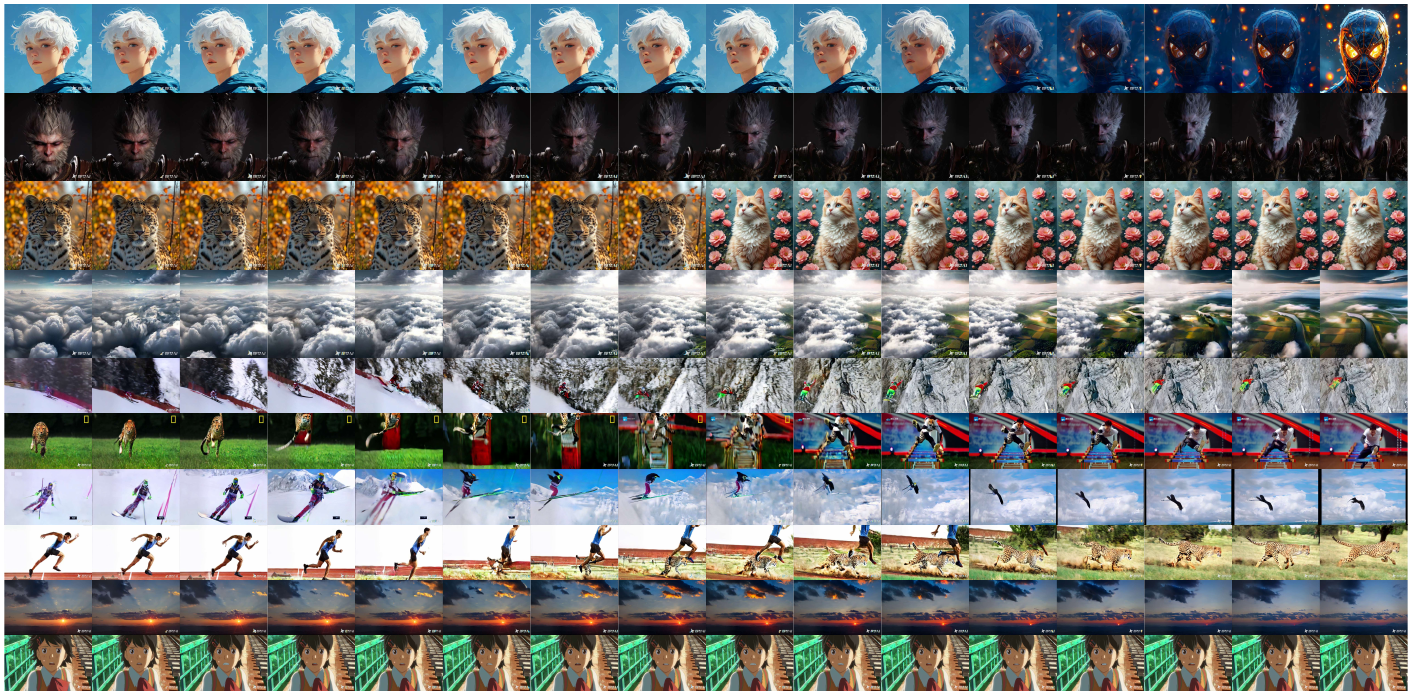}
    \caption{Generated Samples from Jimeng AI}
    \label{fig:jimeng}
\end{figure*}

To further demonstrate the effectiveness of our algorithm, we compared it with several state-of-the-art commercial products, including LUMA AI~\cite{lumalabs}, KLing AI~\cite{keling}, and Jimeng AI~\cite{jimeng}. Due to the varying video generation lengths of these commercial models, direct quantitative comparisons are challenging and may not be fair. Therefore, we conducted a visual comparison analysis instead. Figure~\ref{fig:ours}, Figure~\ref{fig:luma},Figure~\ref{fig:kling} and Figure~\ref{fig:jimeng} present the results of our method compared with these commercial solutions. It is worth noting that LUMA AI and KLing AI typically generate longer videos with fixed settings, which cannot be modified. Since existing open-source models struggle to generate long sequences, we selected 16-frame segments from the commercial products that involve transition content for comparison with our method. For more comprehensive insights, the actual videos can be found in the supplementary multimedia materials. A comparative analysis of these figures reveals that our method consistently produces videos with smooth transitions, whereas the commercial products occasionally fail under varying data conditions. However, commercial products tend to generate richer details, likely because of their underlying robust models. We believe that integrating our approach into these commercial models could potentially enhance their transition generation success rate, resulting in even more seamless and refined video transitions.

\begin{figure*}[htbp]
    \centering
    \includegraphics[width=2\columnwidth]{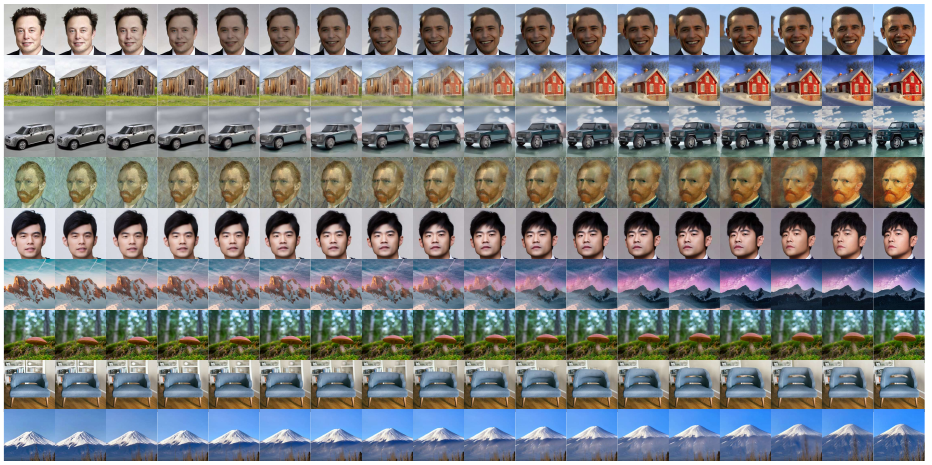}
    \caption{Our Generated Samples from MorphBench Dataset}
    \label{fig:morphbench}
\end{figure*}

Following the quantitative analysis of the TC-Bench-I2V and MorphBench datasets presented in the main text, we now provide additional visualization results generated by our model on these datasets. Figure~\ref{fig:morphbench} and Figure~\ref{fig:tc-bench_results} presents some of these generated examples. Some videos can be viewed in the supplementary multimedia materials. Due to file size limitations for attachments, only a subset of the samples has been included.

\begin{figure*}[htbp]
    \centering
    \includegraphics[width=2\columnwidth]{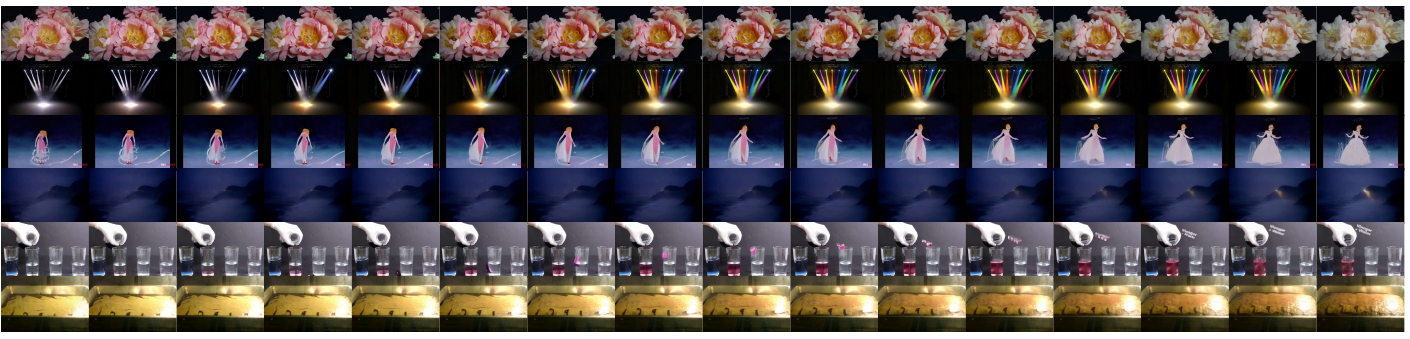}
    \caption{Our Generated Samples from TC-Bench-I2V Dataset}
    \label{fig:tc-bench_results}
\end{figure*}

\end{document}